\documentclass[a4paper,10pt]{article}

\usepackage[left=2.3cm,right=2.3cm,top=2.5cm,bottom=2.5cm]{geometry}
\usepackage[T1]{fontenc}
\usepackage[utf8]{inputenc}
\usepackage{amsmath}
\usepackage{newtxtext}
\usepackage{newtxmath}
\usepackage{graphicx}
\usepackage{booktabs}
\usepackage{tabularx}
\usepackage{array}
\usepackage{multirow}
\usepackage{float}
\usepackage{caption}
\usepackage{xcolor}
\usepackage{enumitem}
\usepackage{algorithm}
\usepackage{algpseudocode}
\usepackage{microtype}
\usepackage{hyperref}
\usepackage{url}
\usepackage{indentfirst}

\graphicspath{{figures/}}
\hypersetup{
  colorlinks=true,
  linkcolor=blue!55!black,
  citecolor=blue!55!black,
  urlcolor=blue!65!black
}
\renewcommand{\arraystretch}{1.15}
\newcolumntype{Y}{>{\centering\arraybackslash}X}
\newcolumntype{Z}{>{\raggedright\arraybackslash}X}

\newcommand{\stateL}{z}
\newcommand{\obs}{o}
\newcommand{\act}{a}
\newcommand{\A}{\mathcal{A}}
\newcommand{\K}{\mathcal{K}}
\newcommand{\C}{\mathcal{C}}
\DeclareMathOperator{\Ret}{Retrieve}
\DeclareMathOperator{\LLMop}{MLLM}
\DeclareMathOperator{\Mem}{Memory}

\title{\textbf{Adaptive Control of Scientific Instruments with Nonstationary Responses through Multimodal Agents}}
\author{
  Guoyang Shi$^{1,2,3}$, Hongwen Li$^{1,2,3}$, Yiding Yu$^{1,2,3}$, Zhuang Fu$^{1,2,3}$, Siqi Ding$^{1,2,3}$,\\
  Zhengyuan Chen$^{1,2,3}$, JianGuo Chen$^{1,2,3}$, Hanyue Zhao$^{1,2,3}$, Tianyuan Liu$^{1,2,3,*}$
}
\date{}

\begin{document}
\maketitle

\begin{center}
\small
$^1$Beijing ENN Fusion Energy Science and Technology Co., Ltd., Beijing 101111, China\\
$^2$Beijing Key Laboratory of High Magnetic Field Spherical Torus Fusion Energy, Beijing 101111, China\\
$^3$Hebei Key Laboratory of Compact Fusion, Langfang 065000, China\\
$^*$Corresponding author: \href{mailto:liutianyuan@enn.cn}{liutianyuan@enn.cn}
\end{center}
\begin{abstract}
\noindent
Controlling scientific instruments requires parameters to be adjusted progressively in response to feedback from successive operations. When unobserved operating conditions evolve with operational history, responses become nonstationary, and a static action--response mapping becomes inadequate for decision-making across shots. Shot-to-shot tuning of neutral beam injection (NBI) ion sources in magnetic-confinement fusion devices is representative of this class of problems: electrode-surface and vacuum conditions are difficult to observe directly, discharge outcomes vary between shots, and experiments on the physical system are costly and carry operational risks. We propose a physics-grounded multimodal agent organized into three independently auditable components. Deterministic algorithms extract symbolic physical quantities and visual evidence from multichannel waveforms; a dual-layer memory organizes within-task trajectories, cross-task experience, and procedural knowledge to support candidate-action generation by a multimodal large language model (MLLM); and an independent action feasibility check enforces action constraints before execution. The MLLM thus integrates evidence and proposes actions, while measurement, action authorization, and final protection remain with deterministic modules. In a strictly chronological replay of 455 shots from four EXL-50U platforms, physical symbols alone outperformed waveform images alone by 0.21--0.32 in quality-classification macro-F$_1$ across all four evaluated models. Retrieved context increased the mean agreement with historical action directions by 22.4--42.7 percentage points, whereas the contribution of visual evidence was model-dependent. Under supervised operation on five ion sources, integrating parameter write-back, programmable logic controller (PLC) execution, and waveform read-back, the system achieved a beam-extraction rate of 86.80\% across 409 executed shots. These results demonstrate the engineering feasibility of the framework for supervised sequential tuning on a physical scientific instrument and provide a basis for evaluating autonomous closed-loop operation and cross-device generalization.

\smallskip
\noindent\textbf{Keywords:}\;neutral beam injection; ion-source tuning; multimodal agent; physical-quantity grounding; retrieval-augmented generation; fusion-device control
\end{abstract}

\section{Introduction}

Frontier scientific discovery increasingly relies on complex scientific instruments to create extreme experimental conditions and acquire high-quality, reproducible observations. As these instruments grow in scale and control dimensionality, bringing them efficiently into reliable operation and making effective use of limited experimental opportunities have become critical to converting major experimental infrastructure into scientific output. Autonomous experimentation offers a potential route by organizing measurement, decision-making, and execution into iterative closed loops \cite{ref1,ref2}. In physical experiments, however, aleatoric and epistemic uncertainties continue to limit the reproducibility and robust operation of closed-loop systems \cite{ref3}. Reliable decisions across successive operations become particularly difficult when the instrument response depends on operational history, important operating conditions are not directly observable, and no high-fidelity simulator is available for repeatable interaction.

Nuclear fusion is regarded as an important pathway toward meeting long-term demand for clean and safe energy \cite{ref4}. As magnetic-confinement fusion devices progress toward high-performance, long-pulse operation, stable auxiliary heating and current drive are essential for plasma heating, confinement, and scenario sustainment \cite{ref5,ref6,ref7,ref8,ref9}. Neutral beam injection (NBI) can deliver megawatt-level heating power, while its beam energy and injection direction can be used to regulate energy deposition and momentum input. It is therefore one of the principal auxiliary-heating techniques for tokamaks and other magnetic-confinement fusion devices \cite{ref10,ref8,ref11}. Before operation, an NBI ion source and its high-voltage acceleration system generally require repeated tuning: an experimental operator sets the discharge and extraction parameters, triggers a discharge, inspects the resulting multichannel waveforms, and then adjusts the parameters for the next shot. Tuning a large NBI ion source may take weeks to months and involve hundreds to thousands of high-voltage or beam pulses \cite{ref12,ref13,ref14}. This experience-dependent, shot-to-shot tuning process directly affects NBI operational efficiency and is representative of the sequential decision-making problem encountered in scientific instruments.

\begin{figure}[H]
  \centering
  \includegraphics[width=0.98\linewidth]{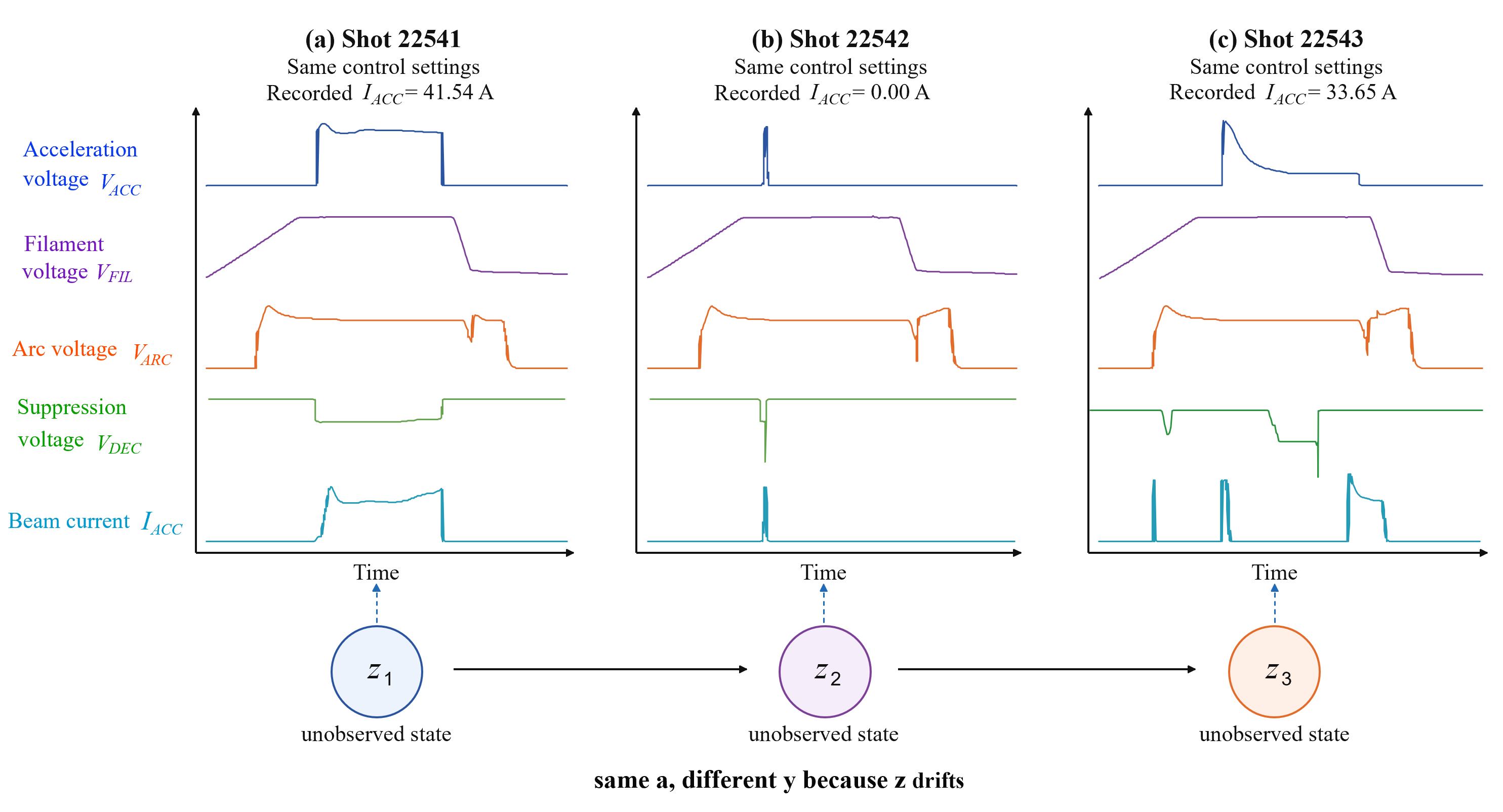}
  \caption{Measured multichannel waveforms from three consecutive shots (shots 22541--22543) under identical control settings. Although all five control variables were held fixed, the recorded acceleration currents were 41.54~A, 0.00~A, and 33.65~A, respectively. The lower states $\stateL_1$--$\stateL_3$ schematically represent unobserved operating conditions and denote history-dependent factors, beyond the current control variables, that may affect the response; the bottom annotation summarizes that the same action $\act$ yields different responses $y$ as the latent state $\stateL$ drifts.}
  \label{fig:latent-drift}
\end{figure}

Ion-source tuning is not a routine parameter-optimization problem defined by a fixed mapping from parameters to performance. Important operating conditions, including electrode-surface state, vacuum cleanliness, and coating condition, are difficult to measure directly and cannot be reset in a controlled manner between shots. Consequently, measured outcomes may vary from shot to shot under identical control settings, as shown in Fig.~\ref{fig:latent-drift}. Moreover, no high-fidelity simulator currently provides a reliable account of shot-to-shot response changes, while exploration on the physical system is costly and risky. The task is therefore more appropriately formulated as a constrained sequential decision-making problem with history-dependent response variation and incomplete observations.

Machine learning has been applied in fusion devices to tasks including plasma-configuration control, confinement-regime identification, and disruption prediction \cite{ref15,ref16,ref17}. Within NBI research, neural networks have been used to accelerate NUBEAM and related NBI-effect and transport simulations \cite{ref18,ref19}; surrogate models and evolutionary algorithms, such as genetic algorithms (GA) and particle swarm optimization (PSO), have been used to learn relationships between ion-source parameters and performance \cite{ref20,ref21,ref22,ref23,ref24,ref25}; and neural networks have also been investigated for predicting ion-source breakdown \cite{ref26}. These studies indicate the potential of simulation-driven control and historical-data modeling for fusion devices and ion-source optimization.

For the ion-source tuning task considered here, these approaches face distinct challenges under the presently available data and device conditions. Reinforcement learning (RL) generally requires either a high-fidelity simulator or extensive interaction with the physical system. Existing models do not yet reproduce shot-to-shot response changes with sufficient accuracy, whereas trial-and-error operation on the physical system is costly and risky. Standard Bayesian optimization is suitable for parameter optimization with expensive observations, and safe variants can provide conditional safety guarantees; however, when the unobserved device state changes across shots, standard or safe Bayesian optimization alone is still insufficient to address latent-state drift directly \cite{ref27,ref28}. Static supervised models that predict outcomes from only the current parameters are poorly suited to exploiting complete tuning trajectories, waveform morphology, and unstructured operating procedures. These limitations point to the same set of missing capabilities: a representation of state across shots, conditioning on operational history, and explicit action constraints before execution.

When neither a suitable simulator nor direct estimation of the latent state is available, a practical alternative is to avoid explicitly reconstructing that state. Instead, the current decision can be conditioned on the recent trajectory of actions and outcomes, retaining traceable cross-shot context without relying on a difficult-to-validate state estimate. Large language model agents have begun to be applied to robotic decision-making \cite{ref29,ref30} and can organize iterative scientific experimental workflows through knowledge retrieval and tool use \cite{ref31,ref32}. Retrieval-augmented generation (RAG) and episodic-memory methods provide traceable interfaces for integrating evolving procedures and long-horizon experience \cite{ref33,ref34,ref35,ref36}. However, observations from scientific instruments include high-frequency time series, waveform images, and equipment states, and time-series data commonly require joint representation with other modalities \cite{ref37}. This work therefore separates deterministic measurement from multimodal reasoning, rather than relying on a general-purpose model to replace quantitative readout. Likewise, semantic guardrails or external validators \cite{ref38} and model-based shielding \cite{ref39} can constrain agent behavior, but they are not equivalent to a deterministic action-membership check against device-permitted bounds.

We propose a physics-grounded multimodal agent for NBI ion-source tuning. The system first uses deterministic algorithms to extract verifiable physical quantities from raw waveforms and anchors waveform images with those quantities. It then generates a candidate action for the next shot by jointly using the recent trajectory of the current task, historical experience, and operating procedures. Finally, before parameters are written, an independent action feasibility check evaluates parameter bounds, per-shot rates of change, and documented exclusion regions. Evidence integration and action proposal are thus assigned to the MLLM, and parameter execution and equipment protection remain with the PLC and hardware interlocks.

Table~\ref{tab:positioning} compares representative technical paradigms in terms of the capabilities required by the problem studied here. ``Handling latent-state drift'' specifically refers to whether a method uses cross-shot history or current context at decision time to adapt to unobserved changes in the device. ``Local--global feature integration'' indicates whether a method can jointly use precisely quantified local features and macroscopic information reflecting the overall state, task objective, or system-wide relationships, and can relate the two during decision-making. ``Action-constraint mechanism'' indicates whether encoded conditions can be checked independently before a candidate action is issued. These dimensions characterize the cited representative studies of each paradigm under the deployment conditions of this study.

\begin{table}[H]
  \centering
  \caption{Qualitative comparison of the task conditions in this study with representative automated decision-making paradigms}
  \label{tab:positioning}
  \small
  \setlength{\tabcolsep}{1.5pt}
  \renewcommand{\arraystretch}{1.18}
  \begin{tabularx}{\linewidth}{|>{\raggedright\arraybackslash}p{2.55cm}|*{6}{Z|}}
    \hline
    Method &
    \centering\shortstack{Handles latent-\\state drift}\arraybackslash &
    \centering\shortstack{Requires a high-\\fidelity simulator}\arraybackslash &
    \centering\shortstack{Uses unstructured\\knowledge}\arraybackslash &
    \centering\shortstack{Local--global\\feature integration}\arraybackslash &
    \centering\shortstack{Action-constraint\\mechanism}\arraybackslash &
    \centering\shortstack{Physical-system\\closed loop}\arraybackslash \\
    \hline
    Ion-source/NBI static surrogate models and GA/PSO \cite{ref20,ref22,ref23,ref24,ref25}
      & Usually no & No & No & Scalar features; no explicit integration
      & Static bounds and constraints can be encoded
      & No reported NBI tuning closed loop \\ \hline
    Simulator-driven plasma RL \cite{ref17}
      & Indirectly, through state observations & Yes & No
      & Local magnetic measurements and global shape objectives
      & Action bounds and device interlocks & Yes (plasma control) \\ \hline
    Standard/safe Bayesian optimization \cite{ref27,ref28}
      & Not addressed for latent-state drift & No & No
      & Point observations and a global surrogate
      & Conditional guarantees in safe variants
      & Applicable to online experiments \\ \hline
    LLM agents for scientific experiments \cite{ref31,ref32}
      & Through context or memory & No & Yes
      & Task-relevant inputs; no explicit integration
      & Can use semantic guardrails or external validators \cite{ref38}
      & Yes (chemical experiments) \\ \hline
    \textbf{This work}
      & \textbf{History-conditioned dual-layer memory} & \textbf{No} & \textbf{Yes}
      & \textbf{Quantitative details and overall waveform morphology}
      & \textbf{Deterministic action feasibility check}
      & \textbf{Supervised shot-to-shot execution} \\ \hline
  \end{tabularx}
\end{table}

The main contributions of this work are as follows:
\begin{enumerate}[leftmargin=2.5em,itemsep=2pt]
  \item \textbf{Problem formulation and layered decision architecture.} We formulate NBI ion-source tuning as a constrained sequential decision-making problem with no available high-fidelity simulator and a partially observed latent device state (Section~\ref{sec:formulation}). On this basis, we delineate the responsibilities of deterministic measurement, evidence retrieval, strategy proposal, and the action feasibility check, such that quantitative readout and execution constraints are handled by deterministic modules and the MLLM concentrates on evidence integration and strategy proposal (Sections~\ref{sec:method} and \ref{sec:safety}).
  \item \textbf{Physics-grounded integration of multimodal evidence and temporal context.} Physical quantities extracted by deterministic algorithms are used to anchor waveform images, which the MLLM interprets jointly with the current control parameters. A dual-layer memory comprising recent within-task trajectories, cross-task historical cases, and procedures then provides traceable temporal context for shot-to-shot decisions under latent-state drift (Sections~\ref{sec:grounding}, \ref{sec:vision}, and \ref{sec:rag}).
  \item \textbf{Cross-model evaluation and execution on physical systems.} We establish a strictly chronological replay protocol covering 455 shots from four EXL-50U~\cite{ref40,ref41,ref42} ion-source platforms. Four foundation models and five evidence configurations are compared to assess the respective contributions of physical symbols, visual evidence, retrieved context, and human feedback. In parallel, parameter write-back, PLC execution, waveform read-back, and next-shot decision-making are integrated on five ion sources, and supervised shot-to-shot execution is used to verify the system's device-integration capability (Sections~\ref{sec:offline-eval} and \ref{sec:online}).
\end{enumerate}

\section{Problem Definition}

\subsection{Ion-Source Physics and Tuning Objectives}

\begin{figure}[H]
  \centering
  \includegraphics[width=0.82\linewidth]{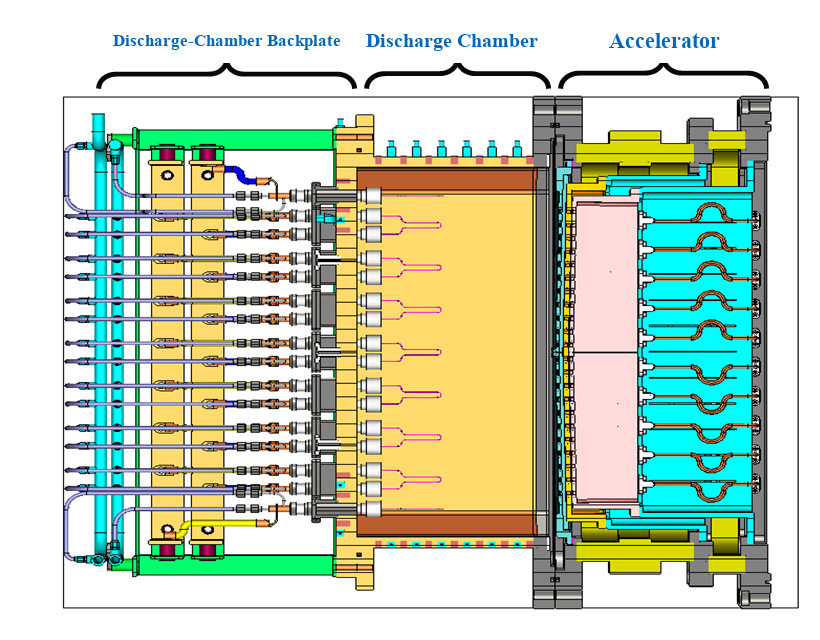}
  \caption{Structure of the 50~kV hot-cathode arc-discharge positive ion source investigated in this study. The device consists primarily of a discharge-chamber backplate integrating the filaments and feedthrough assemblies, a multicusp-field discharge chamber in which the plasma is generated, and a three-electrode accelerator comprising the plasma, suppression, and ground grids.}
  \label{fig:ion-source-structure}
\end{figure}

Beam formation and injection in an NBI system can be divided into four stages: ion production, electrostatic extraction and acceleration, gas neutralization, and neutral-beam injection into the plasma \cite{ref43,ref44}. This study focuses on ion production and electrostatic extraction and acceleration. The ion source under study uses a hot-cathode arc discharge and a multicusp magnetic field, and consists primarily of a discharge chamber and a three-electrode accelerator, as shown in Fig.~\ref{fig:ion-source-structure}. The three electrodes are, in sequence, the plasma grid, suppression grid, and ground grid \cite{ref45}. Heated filaments emit primary electrons, which are accelerated by the arc voltage and ionize the gas through collisions, thereby forming a plasma in the discharge chamber. By suppressing electron losses to the chamber wall, the multicusp magnetic field extends the electron confinement time and affects both the density and spatial uniformity of the plasma. The resulting ions are subsequently extracted, accelerated, and focused through multi-aperture grids.

Shot-to-shot tuning involves five principal control variables---acceleration voltage, filament voltage, arc voltage, gas flow rate, and suppression voltage---which correspond to mutually coupled physical processes. The acceleration voltage primarily determines the energy of individual ions, but the beam current is also jointly constrained by the ion supply and the transport capacity of the accelerator. The filament voltage affects thermionic electron emission; the arc voltage affects the energy of primary electrons and the arc-discharge intensity; and the gas flow rate affects plasma production through the discharge pressure and may also alter collisions, beam loss, and breakdown conditions in the acceleration region. The suppression voltage is used to suppress secondary and backstreaming electrons. An increase in gas pressure or arc current may also increase the suppression-grid current and its ratio to the acceleration current \cite{ref43,ref45}. The effect of any single parameter therefore depends on the other settings and the current device state, and tuning cannot be reduced to a set of independent univariate adjustments.

Ion-source tuning aims to progressively improve discharge stability and the high-voltage hold-off capability of the accelerator. During repeated discharges, quantities such as the microscopic condition of electrode surfaces, adsorbed gas, and vacuum conditions may continue to evolve. Gas breakdown, surface flashover, beam interception, secondary-electron emission, and backstreaming-electron bombardment can induce electrode overcurrent, beam oscillation, or arcing. These factors cannot be measured comprehensively between shots but can affect subsequent discharge outcomes. For conciseness, they are collectively termed the ``latent device state'' in this work. We assess the operational quality of the discharge and extraction processes using voltage, current, and other discharge waveforms. Beam-optics quantities, including spatial uniformity, divergence angle, neutralization efficiency, and the neutral-beam power injected into the plasma, are characterized by dedicated beam diagnostics and can be incorporated as additional tuning objectives in future work.

For each shot, the operator sets the above control variables and evaluates the outcome from multichannel voltage, current, and beam waveforms. The tuning objective is not simply to maximize the acceleration-current amplitude. Drawing on engineering requirements for heating capability, breakdown risk, and device constraints in NBI operation \cite{ref11,ref26,ref9}, this study adopts, subject to device operating limits and constraints on parameter rates of change, the following task objective for the present device: bring the mean acceleration voltage, acceleration-channel electrical power, and effective pulse width to specified targets while avoiding arcing and pronounced oscillations. Based on these physical processes, parameter couplings, and observation boundaries, the device state, control action, measurement outcome, and policy observation are formalized below. A target set $\mathcal{G}$ is used to describe the required operating conditions in a unified manner.

\subsection{Sequential Decision-Making Formulation}\label{sec:formulation}

Consider shot-to-shot tuning of an ion source. Shot $t$ is operated under control action $\act_t$ and produces multichannel measurements, from which the system determines the control action for shot $t+1$. The relevant variables are defined as follows:

\begin{itemize}[leftmargin=2em,itemsep=1pt]
  \item $\act_t \in \A \subset \mathbb{R}^{5}$ is the control action, namely the five-dimensional parameter vector generated and evaluated by the system:
  \begin{equation}
    \act_t =
    \bigl[V_{\mathrm{ACC},t},\,V_{\mathrm{FIL},t},\,V_{\mathrm{ARC},t},\,Q_t,\,V_{\mathrm{DEC},t}\bigr]^{\mathsf T}.
  \end{equation}
  Its five components are, respectively, the acceleration voltage, filament voltage, arc voltage, gas flow rate, and suppression voltage. Other quantities involved in tuning, such as pulse settings and arc-initiation timing, remain nearly constant. They are treated as task conditions in the experiments rather than as action dimensions directly adjusted by the agent. A complete list, including units and constraint forms, is provided in Appendix Table~\ref{tab:parameter-list}.
  \item $\stateL_t \in \mathcal{Z}$ is the \textbf{latent device state}, which characterizes the evolution of electrode-surface morphology, vacuum cleanliness, and coating condition and is \textbf{not directly observable}.
  \item $\mathbf{Y}_t\in\mathbb{R}^{C\times T}$ denotes the $C$ synchronized waveform channels read back from the device.
  \item $\hat{s}_t=\Phi(\mathbf{Y}_t)$ is the symbolic state extracted from the multichannel waveforms by deterministic algorithms. The acceleration-current channel is used as the principal channel to locate the flat top and anomalies; other channels are jointly examined over the same time window to confirm events and compute statistics.
  \item $I_t=\operatorname{Render}\!\left(
  \mathbf{Y}_t^{(I_{\mathrm{ACC}})};\hat{s}_t\right)$
  is the acceleration-current waveform image supplied to the MLLM after deterministic event annotations have been overlaid. The variable $c_t$ denotes textual context, including operator instructions and operating constraints. The observation available to the policy is denoted by
  $\obs_t=(\hat{s}_t,I_t,c_t)$.
\end{itemize}

Let $u_t$ denote external interventions or operational events outside the five-dimensional action space, such as opening the chamber or performing maintenance. Single-shot execution and cross-shot evolution are characterized by two unknown distributions:
\begin{equation}
  \mathbf{Y}_t \sim P_{\mathrm{meas}}(\cdot \mid \stateL_t, \act_t),
  \qquad
  \stateL_{t+1} \sim P_{\mathrm{dyn}}(\cdot \mid \stateL_t, \act_t,u_t).
\end{equation}
After obtaining $\obs_t$, the policy generates the action for the next shot from the task objective and the history of tuning parameters and observations through the current shot:
\begin{equation}
  \act_{t+1}=\pi\bigl(\obs_{1:t},\act_{1:t};g\bigr).
\end{equation}
These relations describe the basic temporal structure of shot-to-shot tuning: the current control action produces the observation for the present shot and, together with external operational events, drives the evolution of the latent device state; the system then uses the information accumulated through the current shot to determine the next action.

Ion-source tuning must simultaneously satisfy multiple criteria, including acceleration voltage, power, pulse width, and stability. We use the target vector $g=(V^{*},P^{*},\tau^{*})$ to represent the task-dependent objectives in a unified manner, where $V^{*}$ is the target acceleration voltage, $P^{*}$ is the acceleration-channel electrical power over the flat-top interval, and $\tau^{*}$ is the minimum effective pulse width. Let $\bar V_{\mathrm{ACC}}(\mathbf{Y})$, $\bar I_{\mathrm{ACC}}(\mathbf{Y})$, and $\tau(\mathbf{Y})$ denote, respectively, the mean effective flat-top acceleration voltage, mean acceleration current, and pulse width deterministically computed from the measurements by the grounding operator, and define
$P_{\mathrm{ext}}(\mathbf{Y})=\bar V_{\mathrm{ACC}}(\mathbf{Y})\bar I_{\mathrm{ACC}}(\mathbf{Y})$.
Because $I_{\mathrm{ACC}}$ may include contributions from ion currents intercepted by the suppression and ground grids, as well as collisional ionization in the gaps \cite{ref45}, $P_{\mathrm{ext}}$ is used only as a power metric for comparisons under consistent operating conditions; it does not represent the net ion-beam power downstream of the accelerator. The operational target set is
\begin{equation}
  \begin{split}
  \mathcal{G}(g)=\bigl\{\mathbf{Y}:\;&
  |\bar V_{\mathrm{ACC}}(\mathbf{Y})-V^{*}|\le\delta_V,\quad
  P_{\mathrm{ext}}(\mathbf{Y})\ge P^{*},\quad
  \tau(\mathbf{Y})\ge\tau^{*},\\
  &n_{\mathrm{bd}}(\mathbf{Y})=0,\quad
  b_{\mathrm{osc}}(\mathbf{Y})=0
  \bigr\},
  \end{split}
  \label{eq:target-set}
\end{equation}
where $\delta_V$ is the permitted deviation of the mean acceleration voltage from its target, $n_{\mathrm{bd}}$ is the number of breakdowns, and $b_{\mathrm{osc}}$ is a shot-level indicator of high-frequency oscillation. If a tuning task does not assess a particular performance criterion, the corresponding condition is removed from the target set for that task.

Shot-to-shot tuning can therefore be formulated as a constrained sequential decision-making problem under partial observability. Based on the histories of observations and actions accumulated through shot $t$, the system generates a control recommendation for shot $t+1$. Each control action must satisfy the device operating constraints, while the system response is progressively brought into, and maintained within, the target set $\mathcal{G}(g)$.

\subsection{Problem Characteristics and Methodological Requirements}

This shot-to-shot tuning problem has three defining characteristics. First, no high-fidelity simulator validated against device data is currently available that can jointly predict $P_{\mathrm{meas}}$ and $P_{\mathrm{dyn}}$. Second, sampling on the physical system is costly: under the standard workflow of the device considered here, executing a shot and reading back its outcome typically takes 3--5~min, while an unsuitable action may also cause arcing, electrode overcurrent, or additional recovery procedures. Third, the latent device state $\stateL_t$ is not directly observable and may evolve across shots. Consequently, after marginalizing over this state, the action--response distribution may vary with operational history. Such evolution does not imply monotonic improvement of the device state with shot number; chamber opening, contamination accumulation, or repeated arcing may instead degrade the state. An empirical analysis of outcome inconsistency among neighboring actions is presented in Section~\ref{sec:drift-result}.

These characteristics distinguish the problem from conventional control tasks with fully observed states and inexpensive interactions, and impose three requirements on the method. In the absence of a high-fidelity simulator, parameter issuance should not be determined directly by response predictions that have not been validated against device data. When the latent state is unobserved and evolving, current decisions should be conditioned on cross-shot historical information. Finally, given the cost of physical experiments and the risk of anomalous operation, strategy proposal should be decoupled from the action feasibility check, while retaining human intervention and hardware interlocks.

\section{Method}\label{sec:method}

\begin{figure}[H]
  \centering
  \includegraphics[width=\linewidth]{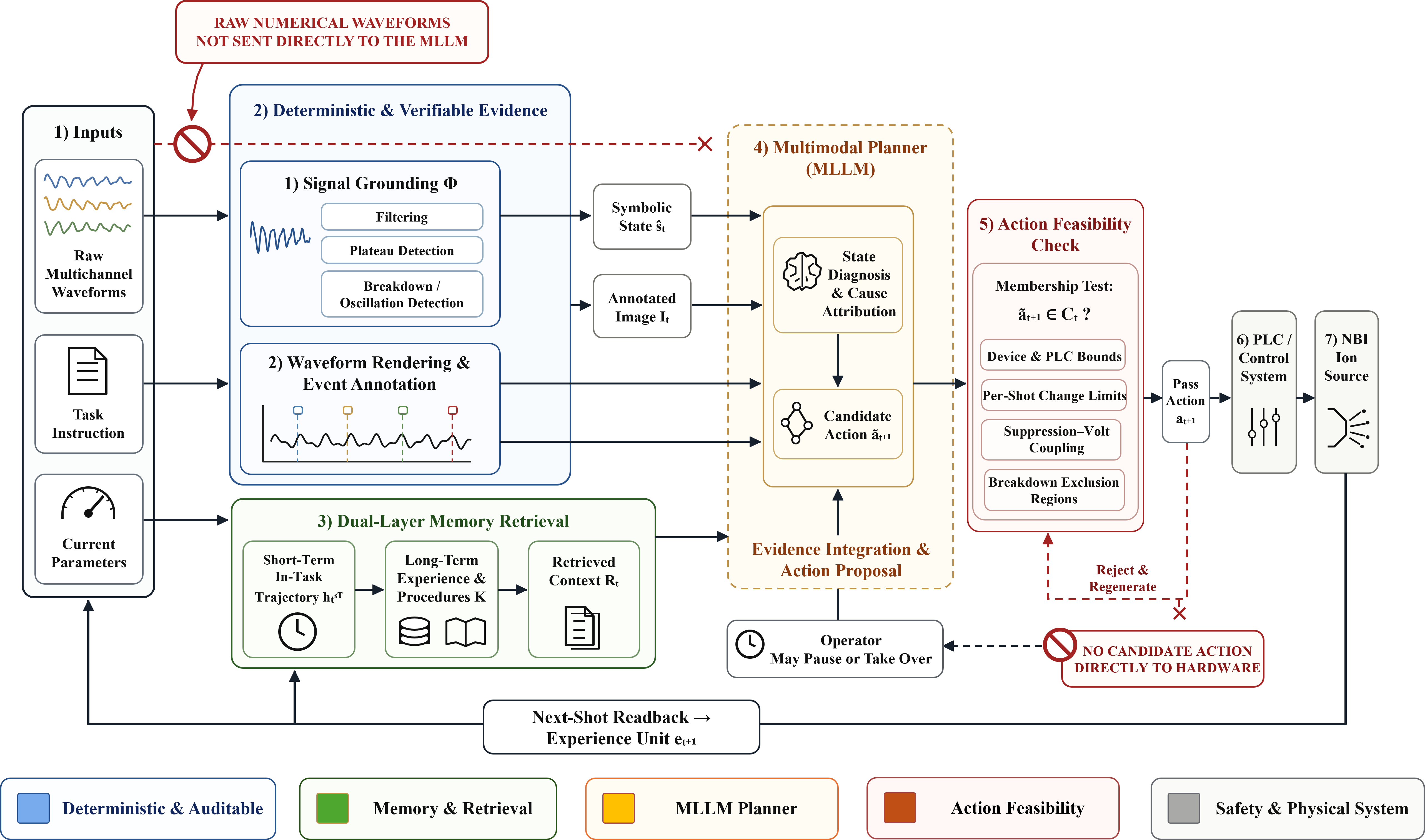}
  \caption{Overall architecture of the physics-grounded multimodal agent. Deterministic signal grounding and annotated waveform rendering transform raw records into symbolic and visual evidence. Dual-layer memory supplies retrieved temporal context, which the MLLM integrates with the current evidence to propose a candidate action for the next shot. The candidate is passed unchanged to the PLC only after an independent feasibility check confirms every condition in $\C_t$; otherwise, it is rejected for regeneration or escalated to the operator. The executed shot is subsequently read back to update the archive and memory.}
  \label{fig:architecture}
\end{figure}

This section presents the implementation of the physics-grounded multimodal agent. In this study, physics grounding refers to evidence represented by deterministic physical-quantity symbols extracted from measurements. The method comprises three core mechanisms: (1) deterministic algorithms extract auditable physical symbols from multichannel waveforms and combine them with visual evidence from the acceleration-current channel to form a multimodal representation; (2) a short-term in-task trajectory is integrated with long-term experience and procedural knowledge to provide traceable context for policy generation under latent-state evolution; and (3) candidate action generation is decoupled from constrained execution, with human intervention triggered under abnormal operating conditions or when the system cannot recover autonomously. The overall system architecture is shown in Fig.~\ref{fig:architecture}, and the complete shot-level decision loop is given in Algorithm~\ref{alg:loop}.

\subsection{Method Overview}

We decompose the overall mapping from the measurement outcome of shot $t$ to the action for shot $t+1$ into five processing stages with clearly delineated responsibilities, each of which can be logged and evaluated independently:
\begin{equation}
  \tilde{\act}_{t+1}
  =\LLMop\bigl(\hat{s}_t,I_t,c_t,\act_t,\mathcal{R}_t,g\bigr),
  \qquad
  \act_{t+1}=\tilde{\act}_{t+1}\;\;\text{if}\;\;
  \tilde{\act}_{t+1}\in\C_t ,
  \label{eq:overall-policy}
\end{equation}
where $\mathcal{R}_t$ denotes the retrieved context jointly formed by the short-term trajectory and long-term experience. The inputs and processing stages are expanded as follows:
\begin{subequations}\label{eq:decomposition}
\begin{align}
  \hat{s}_t &= \Phi(\mathbf{Y}_t)
    && \text{symbolic grounding (Section~\ref{sec:grounding})}, \\
  I_t &= \operatorname{Render}\!\left(
    \mathbf{Y}_t^{(I_{\mathrm{ACC}})};\hat{s}_t\right)
    && \text{image construction (Section~\ref{sec:vision})}, \\
  \mathcal{R}_t &= \Ret\bigl(\hat{s}_t,\act_t,h_t^{\mathrm{ST}},\K\bigr)
    && \text{short-/long-term retrieval (Section~\ref{sec:rag})}, \\
  \tilde{\act}_{t+1}
    &= \LLMop\bigl(
      \hat{s}_t,I_t,c_t,\act_t,\mathcal{R}_t,g
    \bigr)
    && \text{candidate action generation}, \\
  \act_{t+1} &= \tilde{\act}_{t+1},
  \quad \tilde{\act}_{t+1}\in\C_t
    && \text{action feasibility check (Section~\ref{sec:safety})}.
\end{align}
\end{subequations}
Here, $g$ is the objective of the current tuning task and corresponds to Eq.~\eqref{eq:target-set}; $\K$ is the long-term knowledge base containing historical experience and procedural knowledge, and $h_t^{\mathrm{ST}}$ is the short-term trajectory window of the current task. The next-shot candidate action proposed by the planner is denoted by $\tilde{\act}_{t+1}$. The operator $\LLMop$ also generates a structured report containing visual assessments, physical fields, and adjustment rationales. A candidate action is dispatched only after passing the action feasibility check. If it fails, the violated conditions are supplied for regeneration; repeated failures or abnormal operating conditions trigger operator intervention.

\paragraph{Shot-level closed loop.}
After shot $t$, the system retrieves the multichannel waveforms $\mathbf{Y}_t$ and the executed parameters $\act_t$ from MDSplus using the unique shot number, and then constructs the evidence, performs retrieval, generates a candidate action, and conducts the action feasibility check according to Eq.~\eqref{eq:decomposition}. An action that passes the check is written to the PLC. After shot $t+1$, the system reads back the measured outcome and updates the short-term trajectory, thereby forming an ``observation--retrieval--decision--check--execution--feedback'' closed loop.

\subsection{Symbolic Grounding}\label{sec:grounding}

\begin{figure}[H]
  \centering
  \includegraphics[width=\linewidth]{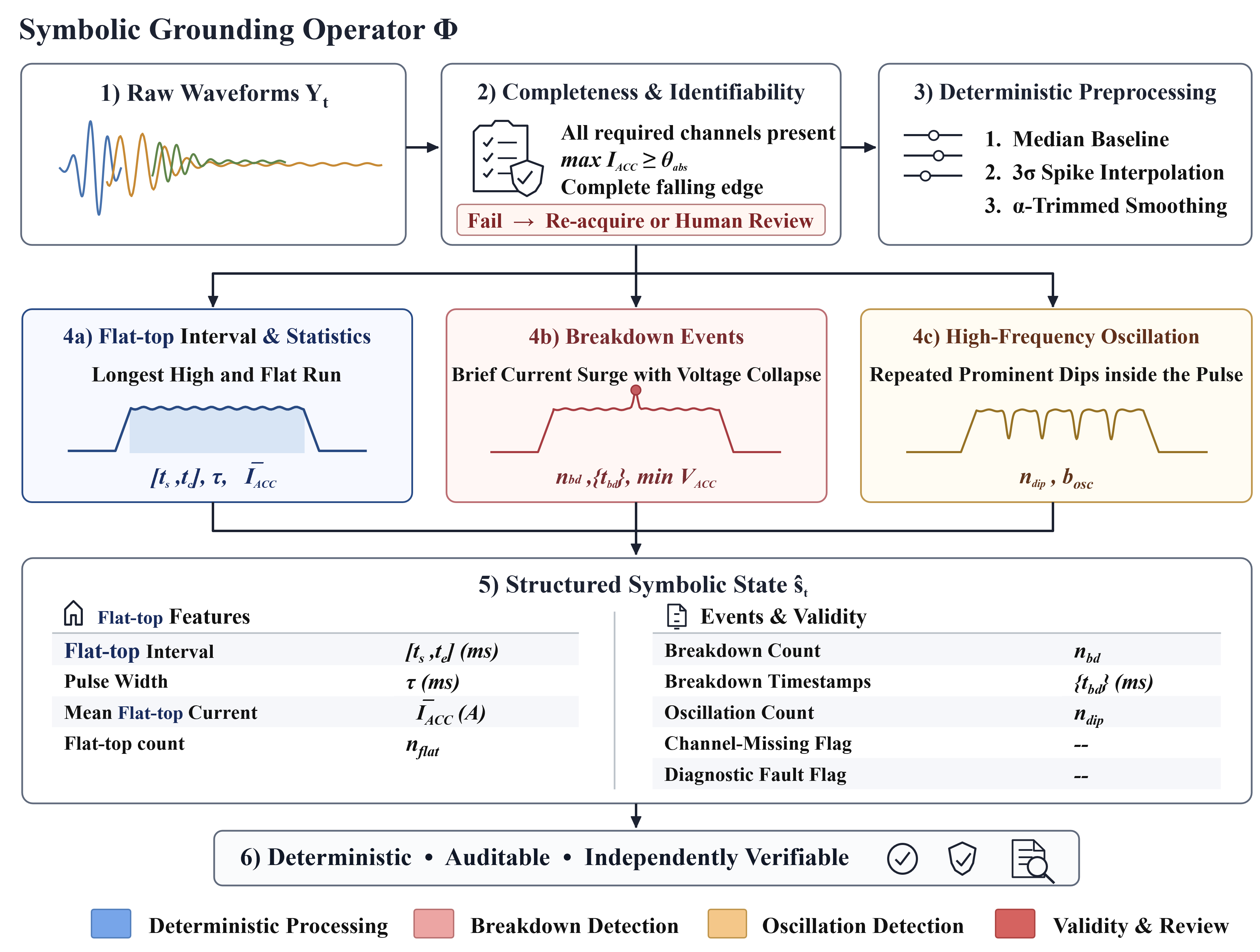}
  \caption{Symbolic grounding pipeline. After preprocessing and interval and event detection, the raw multichannel waveforms yield flat-top statistics, channel-missing flags, breakdown and oscillation events, and fault flags.}
  \label{fig:signal-flow}
\end{figure}

The grounding operator $\Phi$ compiles the multichannel waveforms $\mathbf{Y}_t$ into a symbolic state $\hat{s}_t$. Its outputs include the beam flat-top interval and associated statistics, channel-missing flags, a sequence of breakdown events, an oscillation flag, and fault flags. The breakdown count and shot-level oscillation flag in $\hat{s}_t$ correspond, respectively, to $n_{\mathrm{bd}}(\mathbf{Y}_t)$ and $b_{\mathrm{osc}}(\mathbf{Y}_t)$ in Eq.~\eqref{eq:target-set}. The operator $\Phi$ consists entirely of deterministic algorithms, and its outputs can be audited independently, providing the basis for system auditability. The processing pipeline is shown in Fig.~\ref{fig:signal-flow}.

\subsubsection{Observation Completeness and Identifiability Assessment}
To ensure the auditability of the symbolic state $\hat{s}_t$, the grounding operator $\Phi$ performs quantitative analysis only on observations that satisfy the completeness requirements. A channel $c$ is marked as missing if $\mathbf{Y}_t^{(c)}$ is absent or contains no valid samples. If the acceleration-current channel $I_{\mathrm{ACC}}$ is missing, the flat-top interval, oscillation, and current-related states of the shot are unidentifiable; if the acceleration-voltage channel $V_{\mathrm{ACC}}$ is missing, the breakdown state is unidentifiable. Such observations do not enter the normal decision path described by Eq.~\eqref{eq:overall-policy}, but are instead routed to reacquisition or manual review.

For observations with complete channels, the acceleration-current waveform must also be assessed for suitability for subsequent analysis. If its maximum is below the absolute threshold $\theta_{\mathrm{abs}}$, or if an above-threshold interval extends to the end of the record without a complete falling edge, the waveform diagnostic state is marked as abnormal. This state indicates only that the current waveform does not satisfy the conditions for quantitative analysis; it does not imply a confirmed sensor or hardware failure. Above-threshold segments shorter than the minimum pulse-duration threshold are removed as transient disturbances. If the waveform is valid but contains no pulse satisfying the screening criteria, the outcome is recorded as no valid beam extraction detected rather than as a missing channel.

\subsubsection{Beam Flat-Top Identification}
The beam flat-top interval is the effective beam-extraction interval during which the acceleration current remains relatively stable. Its pulse width and interval statistics are important measures of discharge quality. The algorithm first scans for candidate pulses using the absolute-amplitude threshold $\theta_{\mathrm{abs}}$ and discards overly short segments. It then uses $r_{\mathrm{sel}}=0.8$ times the largest mean current among all candidate pulses as the minimum peak criterion. For any retained pulse, let its acceleration-current sequence be $x_i$. Preprocessing and candidate flat-top determination are expressed as
\begin{equation}
\begin{aligned}
  b_i &= \operatorname{Med}_{w_{\mathrm{med}}}(x)_i,
  & r_i &= x_i-b_i,\qquad \sigma_r=\operatorname{std}(r),\\
  \tilde{x}_i &=
  \begin{cases}
    x_i, & |r_i|\le 3\sigma_r,\\
    \operatorname{Interp}(x)_i, & |r_i|>3\sigma_r,
  \end{cases}\\
  \bar{x}_i &=
  \operatorname{TrimMean}_{\alpha}
  \bigl(\{\tilde{x}_j:j\in\mathcal{W}_i(w_{\mathrm{trim}})\}\bigr),\\
  m_i &= \mathbb{I}\!\left[
    \bar{x}_i\ge r_{\mathrm{flat}}\max_j\bar{x}_j
    \ \land\
    |\nabla\bar{x}_i|\le\theta_{\nabla}
  \right].
\end{aligned}
\end{equation}
Here, $b_i$ is the median-filter baseline, $\operatorname{Interp}$ denotes linear interpolation using adjacent non-spike points, $\mathcal{W}_i$ is a sliding window centered at $i$, and $m_i$ is the candidate flat-top indicator. Within each retained pulse, the algorithm selects the longest contiguous interval for which $m_i=1$ and retains it only if its duration is no less than $\tau_{\min}$. We set $w_{\mathrm{med}}=15$, $w_{\mathrm{trim}}=21$, $\alpha=0.3$, and $r_{\mathrm{flat}}=0.7$. Once the flat-top time window is determined, statistical features of the acceleration-voltage, suppression-grid-current, and other channels are extracted synchronously. All thresholds, units, and sources are listed in Appendix Table~\ref{tab:signal-thresholds}.

\subsubsection{Breakdown Detection}
Breakdowns are generally caused by electrode overcurrent induced by interelectrode gas breakdown, surface flashover, or particle bombardment. In the waveforms, they appear as a brief surge in acceleration current followed by a rapid decrease, accompanied by a drop in acceleration voltage. In this work, ``breakdown'' refers specifically to transient breakdown or overcurrent events and excludes the steady-state suppression-grid current arising from beam divergence, secondary-electron emission, or gap-impact ionization \cite{ref45}. The detection algorithm applies both current and voltage criteria. The current must first reach $I_{\mathrm{hi}}=2.0$~A and then fall below this threshold after remaining high for less than $\Delta t_{\mathrm{bd}}=10$~ms. The program uses the mean acceleration voltage during the first 0.1~s after the start of the record, $V_1$, as the low-level baseline, and sets $V_1+V_{\mathrm{off}}$ ($V_{\mathrm{off}}=5$~kV) as the voltage low-level threshold. Within the window from the onset of the current-threshold crossing to $\Delta t_{\mathrm{off}}=50$~ms after the current falls back, the minimum acceleration voltage must also be below this threshold. Only when both the current and voltage conditions are satisfied does the system identify the event as a breakdown and record its start and end times, duration, minimum voltage, and the time at which that minimum occurs. Here, $V_{\mathrm{off}}$ is the threshold offset relative to the pre-discharge low-level baseline, rather than the magnitude of the operating-voltage drop.

\subsubsection{High-Frequency Oscillation Identification}
\begin{figure}[H]
  \centering
  \includegraphics[width=\linewidth]{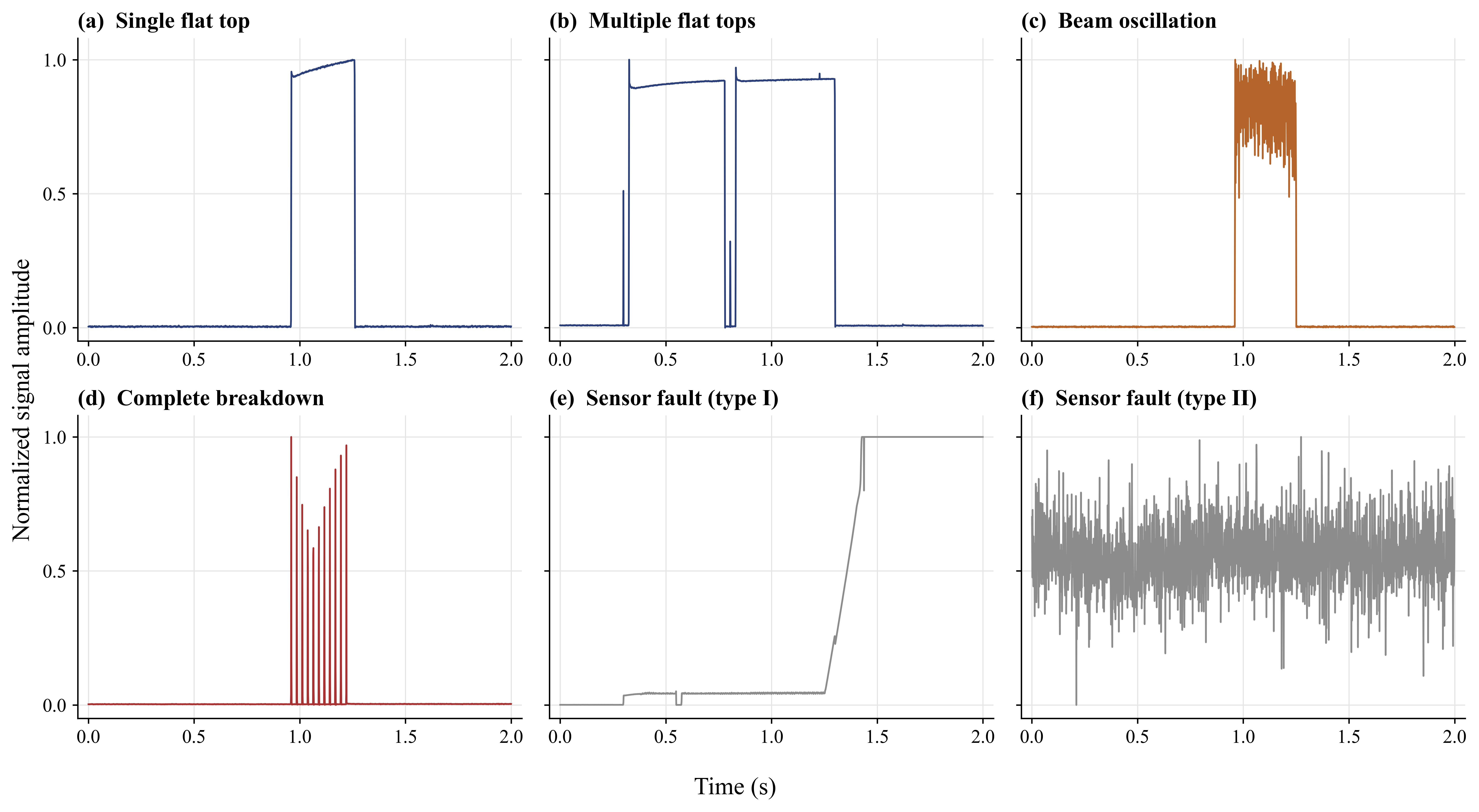}
  \caption{Typical acceleration-current waveform morphologies to be distinguished by the grounding operator: single flat top, multiple flat tops, beam oscillation, complete breakdown, and two types of sensor fault. The vertical axis of each panel is normalized separately and is intended only for morphological comparison.}
  \label{fig:typical-waveforms}
\end{figure}

To detect rapid beam-power drops caused by plasma instability, we use an inverted peak-finding method. The signal is first negated to convert dips into peaks. Small fluctuations are then filtered out using the prominence threshold $\rho_{\mathrm{osc}}$, while sharp dips with peak widths no greater than $w_{\mathrm{osc}}^{\max}=50$ samples are retained without imposing a separate lower bound on peak width. Finally, the depth-variation threshold $\Delta d_{\mathrm{osc}}$ is used to exclude breakdown interference. Let $n_{\mathrm{dip}}$ denote the number of retained prominent dips within the analysis window. The system classifies the waveform as exhibiting high-frequency oscillation when this number reaches $n_{\mathrm{dip}}^{\min}$; here, ``high-frequency'' refers to the repetition rate of dip events per unit time. Figure~\ref{fig:typical-waveforms} shows the typical waveform morphologies that the above algorithms must distinguish.

\subsection{Multimodal Representation with Symbolic Grounding}\label{sec:vision}

\begin{figure}[H]
  \centering
  \includegraphics[width=\linewidth]{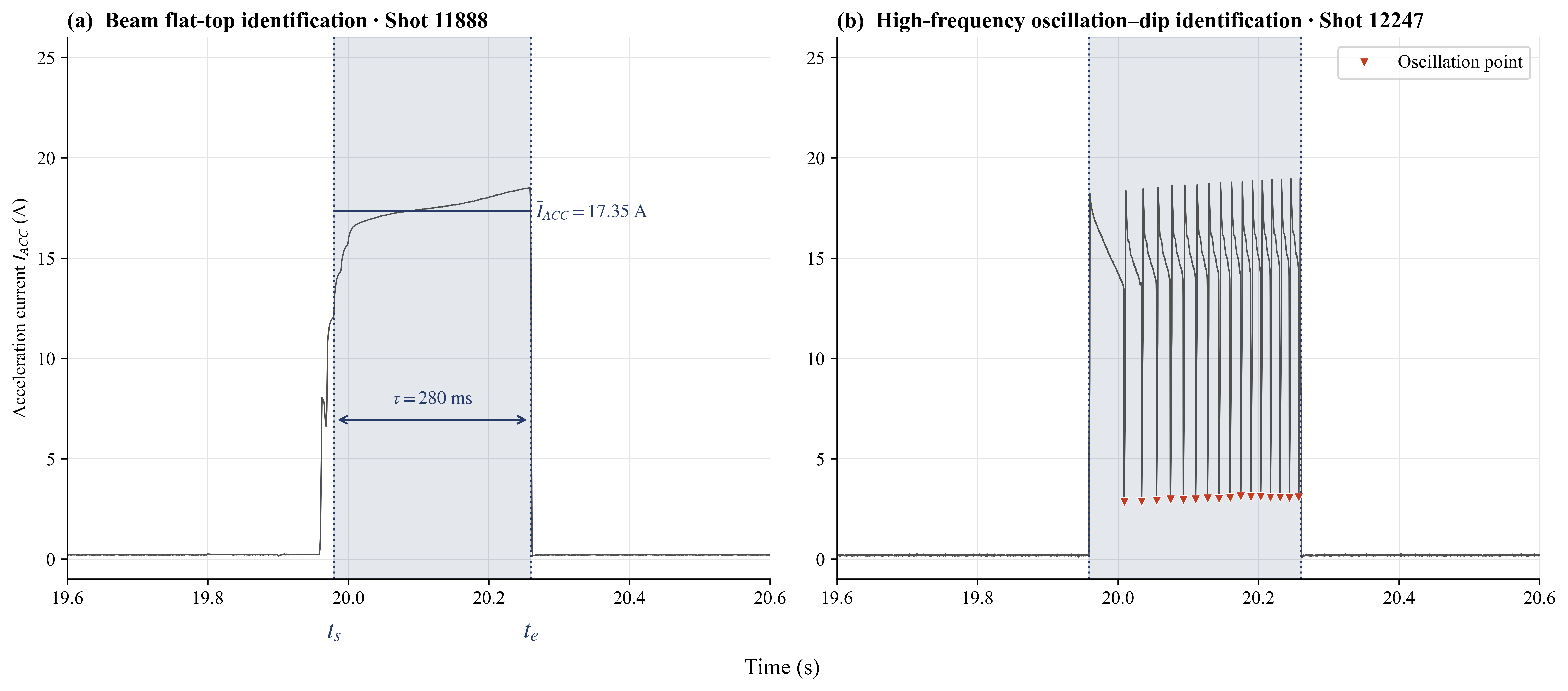}
  \caption{Examples of annotations on measured acceleration-current waveforms: (a) beam flat-top identification for shot 11888; (b) high-frequency oscillation-dip identification for shot 12247. Dark gray denotes the raw waveform; the shaded band bounded by dotted lines marks the detected pulse interval, with the flat-top start $t_s$, end $t_e$, mean current $\bar{I}_{\mathrm{ACC}}$, and width $\tau$ annotated in (a), and red triangles in (b) mark the detected oscillation points.}
  \label{fig:visualization}
\end{figure}

The system renders the acceleration-current waveform as a uniformly formatted annotated image $I_t$ and supplies it directly to the multimodal model for interpretation. The image represents the overall discharge morphology and helps verify whether the physical quantities extracted by $\Phi$ are consistent with the visible structure. To maintain shot-level correspondence between the visual evidence and symbolic state, the raw waveform, $I_t$, and $\hat{s}_t$ are linked by the unique shot number in MDSplus. The symbolic state $\hat{s}_t$ provides quantitative information such as amplitude, pulse width, and flat-top interval, together with state information including channel-missing, breakdown, oscillation, and fault indicators. Before invoking the model, the system performs deterministic checks based on the completeness and fault flags in $\hat{s}_t$. The image $I_t$ enables the multimodal model to identify the overall morphology and locations of local anomalies and assists in cross-checking the physical quantities above; $c_t$ provides the operator instructions and operating constraints for the current shot; and $\act_t$ denotes the control parameters executed for the current shot. These fields are expanded hierarchically when the model is invoked, with $I_t$ supplied as image input. Cross-shot changes are not supplied as a separate field; instead, the model compares the chronologically ordered shot-level records in $h_t^{\mathrm{ST}}$. Consistency between the image and symbolic state is assessed by the multimodal model in context. The model jointly interprets this evidence and generates a candidate action, whose output is subsequently subjected to an independent action feasibility check. Figure~\ref{fig:visualization} provides examples of automatic annotation.

\subsection{Knowledge-Augmented Reasoning and Action Generation}

\subsubsection{Short-Term--Long-Term Dual-Layer Temporal Memory}\label{sec:rag}

Inspired by hybrid multimodal memory and episodic reflection architectures \cite{ref34,ref35}, we employ dual-layer memory: the short-term in-task trajectory memory is denoted by $\Mem_t^{\mathrm{ST}}$, and the long-term experience memory by $\Mem_t^{\mathrm{LT}}$:
\begin{equation}
  \Mem_t=
  \underbrace{\Mem_t^{\mathrm{ST}}}_{\text{short-term in-task trajectory}}
  \cup
  \underbrace{\Mem_t^{\mathrm{LT}}}_{\text{long-term experience and procedural knowledge}}.
\end{equation}
The short-term memory $\Mem_t^{\mathrm{ST}}$ stores the continuous trajectory of the most recent shots in the current task. It is used to compare changes in system response under similar parameter settings, thereby providing historical conditions potentially associated with drift. The long-term memory $\Mem_t^{\mathrm{LT}}$ stores historical experience and procedural knowledge accumulated across tasks and is not discarded as the sliding window of the current task advances. The two memory layers serve distinct roles: the former provides information about state changes within the current task, whereas the latter provides reusable historical experience and procedural knowledge.

\paragraph{Recording and retrieval.}
The latent device state itself is unobservable. The system organizes the recent multi-shot trajectory into experience tuples and writes them directly into the prompt. After shot $t+1$ is completed and its multichannel measurement $\mathbf{Y}_{t+1}$ is read back, the following experience unit is generated:
\begin{equation}
  e_{t+1}=\bigl(\hat{s}_t,\act_{t+1},\hat{s}_{t+1},
  \Delta_{t+1},q_{t+1},\kappa_{t+1}\bigr),
\end{equation}
where ``state change'' refers not to the unobservable latent device state $\stateL$, but to the change in symbolic state produced by the grounding operator. Quantities in $\hat{s}$ used for comparison include the acceleration-current amplitude $\bar I_{\mathrm{ACC}}$, effective pulse width $\tau$, number of flat-top intervals, number of breakdowns, and oscillation flag:
\begin{equation}
  \Delta_{t+1}=\mathcal{D}(\hat{s}_t,\hat{s}_{t+1}).
\end{equation}
Quantitative components are directly subtracted, event variables are represented by differences in counts, and discrete variables record state transitions. Here, $q_{t+1}$ denotes objective-attainment and anomaly labels, while $\kappa_{t+1}$ denotes evidence completeness. Define the ordered history as $h_t=(e_1,\ldots,e_t)$. The short-term trajectory consists of the latest $L$ experience units in this history (we use $L=10$):
\begin{equation}
  h_t^{\mathrm{ST}}
  =\bigl(e_{\max(1,t-L+1)},\ldots,e_t\bigr),
  \qquad
  \Mem_t^{\mathrm{ST}}=h_t^{\mathrm{ST}}.
\end{equation}

When generating the next-shot action, the retrieval operator $\Ret$ uses the current symbolic state and the action executed for the current shot as its query. On the short-term side, it reads all $L$ records in $h_t^{\mathrm{ST}}$ in chronological order; on the long-term side, it retrieves the top-$k$ entries (we use $k=5$) from the knowledge base $\K$ by relevance. Thus, $\mathcal{R}_t=\Ret(\hat{s}_t,\act_t,h_t^{\mathrm{ST}},\K)$, where $\K$ is the retrievable implementation of $\Mem_t^{\mathrm{LT}}$. The knowledge base $\K$ comprises three components: tuning manuals and standard operating procedures; historical cases paired according to parameter configurations, waveform features, and outcome evaluations; and expert rules compiled from structured interviews. Documents are partitioned into semantic chunks according to heading hierarchy and paragraph structure, encoded by an embedding model, and stored in a vector database. Historical cases are stored as complete experience units $e_i$ and associated with ion-source identifiers and provenance information. When generating the next-shot action, the retriever recalls relevant semantic chunks and historical experience units according to the current query and supplies them to the planner as contextual input.

\paragraph{Nonparametric empirical prior.}
In the absence of a high-fidelity cross-shot simulator validated against device data, retrieval provides historical actions and outcomes associated with symbolic states similar to the current one, thereby serving as a nonparametric empirical prior for the candidate policy. This prior requires no gradient update, and new cases can be added to the knowledge base after review. Its effectiveness grows with historical coverage and is maintained through strict causal temporal isolation.

\subsubsection{Multimodal Reasoning and Action Generation}

The multimodal planner receives the current symbolic state, waveform image, textual context, executed control parameters, retrieved context, and task objective. It generates a structured report at three levels: state diagnosis, assessment of possible causes, and policy planning. The report includes diagnostic labels, evidence citations, a five-dimensional candidate action, an adjustment rationale, and an indicative risk level. The risk level is displayed to the operator as an interpretive summary of the model's assessment, while membership in the feasible set and hardware-interlock decisions continue to be determined by the encoded constraints. The system stores the model input, structured output, and associated evidence citations. The measured outcome of the next shot is written to the short-term in-task trajectory, whereas the long-term knowledge base is not updated automatically from model outputs.

Within the predefined task objective and device operating boundaries, a candidate action is written to the PLC only after passing an independent action feasibility check. If a critical channel is missing, the current waveform state is unidentifiable, the feasible set is empty, or the check fails, the system refuses dispatch and requests operator intervention. The operator retains the authority to pause and take over during operation. We adopt supervised shot-to-shot execution in which the operator continuously monitors the process but does not approve every shot; the evaluation protocol on the physical device is described in Section~\ref{sec:online}.

\subsection{Action Feasibility Constraints, Control Execution, and Feedback}\label{sec:safety}

\subsubsection{Action Feasible Set and Check}

Before being written to the hardware, a candidate action must pass a mandatory action feasibility check covering device-permitted boundaries, per-shot change limits, the suppression-voltage coupling relation, and historical breakdown exclusion regions. All relevant voltages are converted to V when calculating the coupling ratio; engineering units and complete constraints are provided in Appendix Table~\ref{tab:parameter-list}. Given the action $\act_t$ executed for shot $t$, the feasible set for the next-shot action is defined as
\begin{equation}
  \C_t =
  \left\{\act\in\mathbb{R}^{5}\;\middle|\;
  \begin{aligned}
    &\act^{\min}\preceq\act\preceq\act^{\max}
      &&\text{(device-permitted boundaries)},\\
    &|\act^{\mathrm{ind}}-\act_t^{\mathrm{ind}}|
      \preceq\delta^{\max}
      &&\text{(per-shot change limits)},\\
    &0.04\,V_{\mathrm{ACC}}\le V_{\mathrm{DEC}}
      \le0.10\,V_{\mathrm{ACC}}
      &&\text{(suppression-voltage coupling)},\\
    &\act\notin\mathcal{F}
      &&\text{(breakdown exclusion regions)}
  \end{aligned}
  \right\},
  \label{eq:feasible}
\end{equation}
where the model generates a five-dimensional candidate action that includes $V_{\mathrm{DEC}}$. The vector $\act^{\mathrm{ind}}=[V_{\mathrm{ACC}},\allowbreak V_{\mathrm{FIL}},\allowbreak V_{\mathrm{ARC}},\allowbreak Q]^{\mathsf T}$ contains the four components subject to independent per-shot change limits, and $\delta^{\max}=[5~\mathrm{kV},\allowbreak 0.5~\mathrm{V},\allowbreak 5~\mathrm{V},\allowbreak 50~\mathrm{sccm}]^{\mathsf T}$ gives their respective limits. By contrast, $V_{\mathrm{DEC}}$ is constrained to the 4\%--10\% coupling interval of $V_{\mathrm{ACC}}$. The bounds $\act^{\min}$ and $\act^{\max}$ are taken from the PLC permission configuration. The set $\mathcal{F}$ comprises high-risk regions summarized by the agent from historical breakdown records and stored with the configuration version; it remains frozen during each shot-level decision.

The action feasibility check does not modify the model-generated candidate action; it performs only a membership check:
\begin{equation}
  \tilde{\act}_{t+1}\in\C_t
  \quad\Longrightarrow\quad
  \act_{t+1}=\tilde{\act}_{t+1}.
  \label{eq:feasibility-check}
\end{equation}
If any condition is violated, the system rejects the candidate and generates the next candidate action with the violated condition attached. At most $n_{\max}=3$ candidate actions, including the initial candidate, are generated and checked for each shot. If all fail, operator intervention is requested. The feasible set is determined by $\act_t$, the current device configuration, and the frozen $\mathcal{F}$, and does not depend on the candidate action. Therefore, if $\C_t=\varnothing$, operator intervention is requested immediately. Human guidance is triggered by a breakdown, repeated generation or check failures, configuration changes, or abnormal operating conditions identified by the operator.

\paragraph{Action-feasibility guarantee.}
Equation~\eqref{eq:feasibility-check} guarantees that only a candidate action that passes the independent membership check can be dispatched; this property does not depend on the correctness of the MLLM. The guarantee covers only the encoded permitted boundaries and historical exclusion regions. It neither guarantees that the subsequent discharge will be free of breakdowns nor covers sensor errors, configuration errors, or unmodeled processes; the hardware interlock retains final shutdown authority.

\subsubsection{Control Execution and Feedback Update}
Parameters that pass the action feasibility check are written to the PLC through a database interface and read and executed by the PLC on the next shot. Using the unique shot number, the system interface associates the configuration, waveforms, retrieved evidence, model report, and check log. Automatic triggering is halted upon a write or communication failure or activation of an interlock. After the discharge, the system reads back the waveforms and writes the experience unit $e_{t+1}$ to the short-term trajectory. The long-term knowledge base is not automatically modified by individual-shot outputs; feedback updates occur only in external memory.

\begin{algorithm}[H]
\caption{Shot-level decision loop of the physics-grounded multimodal agent}
\label{alg:loop}
\begin{algorithmic}[1]
\Require Task objective $g$; textual context $c_t$; short-term trajectory $h_t^{\mathrm{ST}}$; knowledge base $\K$; action $\act_t$ for shot $t$
\Ensure Dispatched action $\act_{t+1}$ for shot $t+1$, or a rejection status
\State $\mathbf{Y}_t \gets$ read multichannel waveforms from MDSplus
\State $\hat{s}_t \gets \Phi(\mathbf{Y}_t)$ \Comment{Deterministic, auditable multichannel symbolic grounding}
\State $I_t \gets$ render the acceleration-current waveform and overlay event annotations from $\hat{s}_t$
\If{$\hat{s}_t$ flags a missing critical channel or a fault that invalidates the current decision}
  \State Request reacquisition or manual review; do not dispatch this automatic action; \textbf{return}
\EndIf
\State $\mathcal{R}_t \gets \Ret(\hat{s}_t,\act_t,h_t^{\mathrm{ST}},\K)$ \Comment{Read the short-term trajectory and retrieve long-term knowledge}
\State $\tilde{\act}_{t+1} \gets
  \LLMop(\hat{s}_t,I_t,c_t,\act_t,\mathcal{R}_t,g)$
\State $\C_t\gets$ construct the feasible set from action bounds, per-shot change limits, the coupling interval, and $\mathcal{F}$
\State $r\gets1$; $b_{\mathrm{valid}}\gets\mathrm{false}$ \Comment{$r$ is the number of candidates generated}
\While{$r\le n_{\max}$ \textbf{and} $b_{\mathrm{valid}}=\mathrm{false}$}
  \If{$\C_t=\varnothing$}
    \State \textbf{break}
  \EndIf
  \If{$\tilde{\act}_{t+1}\in\C_t$}
    \State $b_{\mathrm{valid}}\gets\mathrm{true}$
  \Else
    \If{$r<n_{\max}$}
      \State $r\gets r+1$
      \State $\tilde{\act}_{t+1}\gets$ generate the next candidate with the violated constraints attached
    \Else
      \State \textbf{break}
    \EndIf
  \EndIf
\EndWhile
\If{$b_{\mathrm{valid}}=\mathrm{false}$}
  \State Request operator intervention; do not dispatch this automatic action
  \State \Return
\EndIf
\State $\act_{t+1}\gets\tilde{\act}_{t+1}$
\State Dispatch $\act_{t+1}$; trigger the discharge; acquire $\mathbf{Y}_{t+1}$
\State $\hat{s}_{t+1}\gets\Phi(\mathbf{Y}_{t+1})$
\State Render $I_{t+1}$ from $\hat{s}_{t+1}$ and construct $\obs_{t+1}$
\State Construct $\Delta_{t+1}$, $q_{t+1}$, and $\kappa_{t+1}$ from $\hat{s}_{t+1}$, and generate $e_{t+1}$
\State Write experience unit $e_{t+1}$ to the short-term in-task trajectory
\end{algorithmic}
\end{algorithm}

\section{Experimental Setup}

\subsection{Data and Comparators}
The experiments covered four independent ion-source platforms of the EXL-50U neutral beam injection system. Each platform was equipped with an independent ion source, control system, and data acquisition system, and differed in power-supply characteristics, vacuum conditions, and electrode states. The unified offline evaluation set comprised historical records from 455 shots. Multiple experts jointly reviewed the raw multichannel waveforms and reached a single consensus label through discussion, assigning each sample to one of six classes: full breakdown, insufficient pulse width, long-pulse high current, long-pulse low current, high-frequency oscillation, and fault noise. This discharge-quality taxonomy defines the label space for the classification metrics reported in this study. The platform-specific sample sizes were 127 shots from Test Bench 1, 116 from Test Bench 2, 110 from Experimental Bench 1, and 102 from Experimental Bench 2. Here, ``platform'' denotes a complete ion-source, control, and acquisition system, whereas ``ion source'' denotes the physical ion-generating component within that system. Offline records are identified at the platform level, whereas online records are further identified by the individual ion source within each platform.

The offline experiments employed strict chronological replay advancing shot by shot. When predicting the action for shot $t+1$, the current shot $t$ served as the evaluation input, and the retriever could access only records already available before that decision point. Thus, the experiment evaluates waveform interpretation and next-shot action reproduction along the historical operating sequence of the device, reproducing the information conditions available to operators at each decision point. The on-device experiments used the Qwen3-VL-Plus Full configuration for supervised shot-to-shot execution and engineering integration on the physical system. The on-device summary comprised 409 execution records from five ion sources. Among these, three on-device experiments conducted on the same ion source at Test Bench 2 retained complete raw waveforms for 23 shots and were used for the shot-to-shot analysis in Section~\ref{sec:online}.

\begin{table}[H]
  \centering
  \caption{Distribution of the six waveform classes in the 455-shot offline dataset}
  \label{tab:dataset-example}
  \small
  \begin{tabular}{lr}
    \toprule
    Class & Number of shots \\
    \midrule
    Full breakdown & 67 \\
    Insufficient pulse width & 113 \\
    Long-pulse high current & 80 \\
    Long-pulse low current & 81 \\
    High-frequency oscillation & 51 \\
    Fault noise & 63 \\
    \midrule
    Total & 455 \\
    \bottomrule
  \end{tabular}
\end{table}

Table~\ref{tab:dataset-example} summarizes the six-class consensus labels obtained through direct expert review of the raw multichannel waveforms. The human ground truth was not generated by the grounding operator $\Phi$ or by model outputs. On the model side, the number of flat tops, pulse width, current amplitude, and breakdown flag were read from the structured report generated under each configuration and mapped to the same six classes using a common prespecified rule: $n_{\mathrm{flat}}=0$ with breakdown was classified as full breakdown; $n_{\mathrm{flat}}=0$ without breakdown as fault noise; $n_{\mathrm{flat}}\ge2$ as high-frequency oscillation; and cases with $n_{\mathrm{flat}}=1$ were assigned to the remaining three classes according to pulse width and current amplitude. The criterion $n_{\mathrm{flat}}\ge2$ for high-frequency oscillation was motivated by interruption and subsequent re-establishment of the acceleration current within a single shot, which corresponded to oscillation-induced power drops in this dataset. This criterion uses only the count field in $\hat{s}_t$ and does not invoke the dip detector described in Section~\ref{sec:grounding}. The mapping does not use the human class label of the evaluated shot. Human labels and model predictions were linked by the unique shot identifier before cross-model evaluation.

\subsection{Unified Cross-Model Evaluation Protocol}
The cross-model comparison included Qwen3.5-Plus, MiniMax3, GLM-5.3-Flash, and Qwen3-VL-Plus. Each model was evaluated under five configurations: Baseline, Physical-only, +Physical, +RAG, and Full. Baseline received only the raw waveform images and task text. Physical-only received only the physical symbols extracted by deterministic algorithms and the task text, without waveform images. +Physical received both the physical symbols and waveform images annotated in a standardized format. +RAG additionally incorporated retrieved content from the recent operational trajectory, long-term experience, and procedural knowledge. Full further enabled the human-guidance and feedback protocol for abnormal operating conditions. The configurations are defined in Table~\ref{tab:ablation-settings}.

The five configurations support three groups of comparisons. The comparison between Baseline and +Physical characterizes the overall gain of the complete physical-grounding scheme, which combines physical symbols with annotated images. The comparison between Physical-only and +Physical measures the incremental contribution of visual evidence when physical symbols are available. The successive transitions from +Physical to +RAG and from +RAG to Full measure the contributions of retrieved context and the human-feedback protocol, respectively, with Full representing system performance under the human--machine feedback mode adopted in supervised operation.

\begin{table}[H]
  \centering
  \caption{Five unified configurations for cross-model evaluation}
  \label{tab:ablation-settings}
  \small
  \begin{tabularx}{\linewidth}{|l|Y|Y|Y|Y|Y|}
    \hline
    Configuration & Symbolic grounding & Visual evidence & Experience and knowledge retrieval & Feasibility check & Human feedback \\ \hline
    Baseline  & $\times$     & $\checkmark$ & $\times$     & $\checkmark$ & $\times$ \\ \hline
    Physical-only & $\checkmark$ & $\times$ & $\times$     & $\checkmark$ & $\times$ \\ \hline
    +Physical & $\checkmark$ & $\checkmark$ & $\times$     & $\checkmark$ & $\times$ \\ \hline
    +RAG      & $\checkmark$ & $\checkmark$ & $\checkmark$ & $\checkmark$ & $\times$ \\ \hline
    Full      & $\checkmark$ & $\checkmark$ & $\checkmark$ & $\checkmark$ & $\checkmark$ \\ \hline
  \end{tabularx}
\end{table}

Under a given configuration, all models used exactly the same evaluated shots, task text, prompt templates, decoding constraints, and action bounds. Waveform images, symbolic quantities, retrieval results, and human feedback were enabled or disabled strictly according to Table~\ref{tab:ablation-settings}. When $\obs_t$ was used to generate a recommendation for shot $t+1$, the outcome of shot $t$ was already observable, and the recent trajectory $h_t^{\mathrm{ST}}$ could include the experience unit $e_t$ derived from that shot. The long-term knowledge base, however, was strictly restricted to $\K_{<t}$ and could not access $e_{t+1}$ or any subsequent records, the human class label of the evaluated shot, or retrospective expert feedback. Multiple views and adjacent slices from the same shot were always bound to the same shot identifier. Signal thresholds, prompt templates, retrieval parameters, examples, and action constraints were fixed before comparison. Human feedback in the Full configuration was drawn only from historical operator records already available before the corresponding decision point; it did not use the evaluated-shot outcome, the current human label, or retrospectively generated corrections.

\subsection{Metrics, Implementation, and Statistical Analysis}
The evaluation comprised four levels. First, for quantitative readout, the acceleration-current amplitude, pulse width, and number of flat tops in each structured report were compared with the reference values of the evaluation set using the mean absolute error (MAE) and root-mean-square error (RMSE). Under Baseline, this level measures how accurately the model recovers these quantities from waveform images; under Physical-only and all subsequent configurations, the corresponding quantities extracted by $\Phi$ were included in the prompt, and the same metrics measure how accurately the model reports them when deterministic readings are available. Second, the direct quality-classification field in the model's structured report was evaluated using Accuracy and macro-F$_1$, hereafter termed quality-classification macro-F$_1$. Third, to assess whether the physical fields in the model report support its classification conclusion, a separate six-class prediction was generated from the reported number of flat tops, pulse width, current amplitude, and breakdown flag using the prespecified mapping described in the preceding subsection, and macro-F$_1$ was computed against the same expert consensus labels. This metric is hereafter termed field-derived six-class macro-F$_1$. These are two distinct readouts from the same report and are not interchangeable. Fourth, the flat-top, breakdown, and continuous-value outputs of $\Phi$ were evaluated independently of the large language model. Operator outputs were matched by shot identifier to the human reference labels across the four platforms. We report MAE for acceleration current and main flat-top width, exact-count accuracy for flat tops and breakdowns, and count-based event F$_1$.

Reproduction of expert action directions was evaluated separately for five parameters using three directional classes: increase, hold, and decrease. For parameter $k$ of evaluated shot $t$, the historical expert direction and the system-recommended direction were defined, respectively, as
\begin{equation}
  d^{\mathrm{E}}_{t,k}
  =\operatorname{sgn}\!\left(a^{\mathrm{E}}_{t+1,k}-a_{t,k}\right),
  \qquad
  \hat d_{t,k}
  =\operatorname{sgn}\!\left(\hat a_{t+1,k}-a_{t,k}\right),
\end{equation}
where $a_{t,k}$, $a^{\mathrm{E}}_{t+1,k}$, and $\hat a_{t+1,k}$ denote, respectively, the executed parameter for shot $t$, the parameter actually adopted by the historical expert for the next shot, and the system-recommended parameter, all recorded at the effective PLC control resolution. An increase, exact match, or decrease in the setpoint was labeled as increase, hold, or decrease, respectively. The historical tuning records are dominated by increase and hold actions, consistent with the stepwise parameter advancement that characterizes ion-source conditioning. Directional agreement (Accuracy) was therefore adopted as the primary metric for action reproduction and was reported separately for the five parameters, together with the arithmetic mean of the five directional agreement values. This metric quantifies the extent to which the system reproduces the adjustment directions adopted by experienced operators under the same observations, and thus measures the consistency of its decisions with expert tuning practice. Adjustment magnitude and next-shot physical outcomes are assessed separately in the on-device experiments in Section~\ref{sec:online}.

For $N$ samples,
\begin{align}
  \mathrm{MAE} &= \frac{1}{N}\sum_{i=1}^{N}\left|y_i-\hat{y}_i\right|, &
  \mathrm{RMSE} &= \sqrt{\frac{1}{N}\sum_{i=1}^{N}\left(y_i-\hat{y}_i\right)^2}, \\
  \mathrm{Accuracy} &= \frac{1}{N}\sum_{i=1}^{N}\mathbf{1}[\hat y_i=y_i], &
  F_{1,c} &= \frac{2P_cR_c}{P_c+R_c},\qquad
  F_1^{\mathrm{macro}}=\frac{1}{C}\sum_{c=1}^{C}F_{1,c}.
\end{align}
Here, $C$ is the number of classes. If both precision and recall are zero for a class, $F_{1,c}$ is set to zero. Offline results were obtained from the fixed 455-shot evaluation set; we report the observed metrics and paired within-shot differences on this set. The on-device experiments were summarized according to whether beam extraction occurred, and shot-to-shot analyses were conducted for the three experiments with complete raw waveforms.

MDSplus was used for the data interface, ChromaDB for vector retrieval, and SQLite for structured states and version indices. In the offline cross-model evaluation, waveform processing, image rendering, retrieval, and the action feasibility check used engineering configurations fixed before comparison, and shot-level outputs were retained. Each LLM used a fixed model snapshot and a common structured-output constraint, without training or fine-tuning.
\section{Results and Discussion}

\subsection{Analysis of Nonstationarity in Device Response}\label{sec:drift-result}
The proposed method is motivated by the following premise: if the current control settings were sufficient to determine the next-shot outcome, shot-to-shot tuning could be reduced to fitting a static mapping, obviating the need for history-conditioned memory and shot-level feedback. We therefore first examined whether similar actions consistently corresponded to similar waveform outcomes by comparing action-neighbor shot pairs from the same platform and with the same set pulse width. The action for each shot was represented by the five-dimensional vector
$a=(V_{\mathrm{ACC}},V_{\mathrm{FIL}},V_{\mathrm{ARC}},Q,V_{\mathrm{DEC}})$.
Each component was range-normalized by its PLC-permitted interval\cite{ref46}, and the action distance for shot pair $(i,j)$ was defined using the Chebyshev distance\cite{ref47}:
\begin{equation}
  D_{\act}(i,j)=
  \max_{k\in\{1,\ldots,5\}}
  \frac{|a_{i,k}-a_{j,k}|}{a_k^{\max}-a_k^{\min}},
  \qquad
  D_y(i,j)=\mathbf{1}[\ell_i\ne\ell_j],
\end{equation}
where $a_{i,k}$ is component $k$ of the action vector for shot $i$, and $a_k^{\max}-a_k^{\min}$ is the PLC-permitted range for that component. Division by the corresponding range places control variables with different units on a common relative scale. The outer maximum makes the pairwise distance equal to the largest relative difference among the five components. Here, $\ell_i$ is the expert consensus quality label for shot $i$; $D_y(i,j)=1$ when the two shots have different labels and 0 otherwise. Candidate shot pairs were defined as
\begin{equation}
  \mathcal{P}=\{(i,j):s_i=s_j,\ W_i=W_j,\ i<j\},
\end{equation}
where $s_i$ and $W_i$ denote the platform and set pulse width, respectively. The condition $i<j$ prevents duplicate counting of the same shot pair. Thus, $\mathcal{P}$ retains only comparable pairs from the same platform and with the same set pulse width. The action-neighbor outcome conflict rate on $\mathcal{P}$ was defined as
\begin{equation}
  r_{\mathrm{drift}}=
  \frac{\sum_{(i,j)\in\mathcal{P}}\mathbf{1}[D_{\act}(i,j)\le0.05]D_y(i,j)}
       {\sum_{(i,j)\in\mathcal{P}}\mathbf{1}[D_{\act}(i,j)\le0.05]}.
\end{equation}
The indicator $\mathbf{1}[D_{\act}(i,j)\le0.05]$ equals 1 only for shot pairs satisfying the action-neighbor criterion. The denominator counts all action-neighbor pairs, whereas the numerator additionally uses $D_y(i,j)$ to retain pairs with discordant outcome labels. Their ratio is therefore the proportion of similar actions that produced different quality outcomes. The primary analysis used $D_{\act}\le0.05$, meaning that every action component differed by no more than 5\% of its permitted range. Because individual shots were shared across pairs, shots were resampled with replacement within each $(s,W)$ stratum, and a percentile bootstrap ($B=2000$) was used to estimate the 95\% confidence interval\cite{ref48}. Treating shared shots as the resampling units accounts for dependence among overlapping pairs\cite{ref49}.

At the primary threshold, 2123 neighboring pairs were identified, yielding $r_{\mathrm{drift}}=54.7\%$ (95\% confidence interval, 47.8\%--60.7\%). When the threshold was changed to 0.02 and 0.10, the conflict rates were 55.5\% (1307 pairs) and 58.3\% (3990 pairs), respectively. Even among the 433 pairs with exactly identical parameters, the conflict rate remained 44.1\%. After neighboring pairs were divided at the median inter-shot interval, the conflict rate was 14.5 percentage points higher in the more widely separated group. A permutation test that randomly shuffled labels within each stratum yielded $p<0.001$\cite{ref50}. These results show that similar actions do not necessarily produce similar waveform outcomes and suggest that the system response is nonstationary and related to operating history. This observation provides empirical motivation for the proposed design: shot-to-shot decisions require not only the current observation but also auditable quantitative grounding and history-conditioned context. The following two subsections accordingly evaluate this design on the offline dataset and the physical device, respectively.

\subsection{Offline Cross-Model Evaluation and Module Ablation}\label{sec:offline-eval}
This subsection addresses three questions: whether deterministic physical symbols provide reliable quantitative grounding for waveform interpretation; what incremental contributions visual evidence and retrieval memory make to perception and decision-making, respectively; and whether these observations remain consistent across models. To this end, five configurations of Qwen3.5-Plus, MiniMax3, GLM-5.3-Flash, and Qwen3-VL-Plus were compared on the fixed 455-shot evaluation set. Table~\ref{tab:real-cross-model} reports the overall results. Complete component-wise results for recognition and action-direction reproduction are provided in Appendix Tables~\ref{tab:recognition-detail-all} and~\ref{tab:action-detail-all}.

\paragraph{Validation of the grounding operator.}
We first independently compared the outputs of $\Phi$ against the reference labels for 455 matched shots from the four platforms. As shown in Table~\ref{tab:grounding-validation}, the pooled acceleration-current MAE was 0.68~A and the pooled main flat-top-width MAE was 10.42~ms. Exact-count accuracy and count-based event F$_1$ were 97.14\% and 97.61\%, respectively, for flat tops, and 99.78\% and 99.66\% for breakdowns. Flat-top event-count F$_1$ was at least 96.08\% on every platform, while breakdown event-count F$_1$ was at least 97.56\%. These results show that the deterministic grounding operator consistently extracts the principal quantitative measurements and event counts subsequently used for multimodal reasoning across platforms.

\begin{table}[H]
  \centering
  \caption{Independent validation of the symbolic grounding operator $\Phi$ on four platforms}
  \label{tab:grounding-validation}
  \small
  \setlength{\tabcolsep}{3.5pt}
  \begin{tabular}{@{}l*{7}{c}@{}}
    \toprule
    Platform & Shots &
    \shortstack{Current\\MAE/A} &
    \shortstack{Main-width\\MAE/ms} &
    \shortstack{Flat-top-count\\accuracy} &
    \shortstack{Flat-top event-\\count F$_1$} &
    \shortstack{Breakdown-count\\accuracy} &
    \shortstack{Breakdown event-\\count F$_1$} \\
    \midrule
    Test Bench 1 & 127 & 0.16 & 4.46  & 100.00\% & 100.00\% & 100.00\% & 100.00\% \\
    Test Bench 2 & 116 & 0.64 & 14.35 & 95.69\%  & 96.08\%  & 100.00\% & 100.00\% \\
    Experimental Bench 1 & 110 & 0.70 & 13.24 & 98.18\%  & 97.44\%  & 99.09\%  & 97.56\% \\
    Experimental Bench 2 & 102 & 1.34 & 10.35 & 94.12\%  & 96.20\%  & 100.00\% & 100.00\% \\
    \midrule
    \textbf{Pooled} & \textbf{455} & \textbf{0.68} & \textbf{10.42} &
    \textbf{97.14\%} & \textbf{97.61\%} & \textbf{99.78\%} & \textbf{99.66\%} \\
    \bottomrule
  \end{tabular}
\end{table}

\begin{table}[H]
\centering
\caption{Overall results for four models under five configurations}
\label{tab:real-cross-model}
\small
\setlength{\tabcolsep}{1pt}
\begin{tabularx}{\linewidth}{@{}ll*{5}{>{\centering\arraybackslash}X}@{}}
\toprule
Model & Configuration & \shortstack{Quality-classification\\macro-F$_1$} & \shortstack{Current\\MAE/A} & \shortstack{Pulse-width\\MAE/ms} & \shortstack{Flat-top-count\\MAE} & \shortstack{Mean directional\\agreement} \\
\midrule
Qwen3.5-Plus & Baseline & 0.47 & 3.67 & 138.19 & 0.27 & 36.7\% \\
Qwen3.5-Plus & Physical-only & 0.79 & 1.40 & 28.37 & 0.09 & 54.4\% \\
Qwen3.5-Plus & +Physical & 0.73 & 1.54 & 42.23 & 0.09 & 55.8\% \\
Qwen3.5-Plus & +RAG & 0.71 & 1.24 & 19.36 & 0.09 & 78.2\% \\
Qwen3.5-Plus & Full & 0.68 & 1.31 & 24.33 & 0.08 & 86.0\% \\
\midrule
MiniMax3 & Baseline & 0.57 & 3.91 & 107.80 & 0.24 & 35.6\% \\
MiniMax3 & Physical-only & 0.78 & 2.34 & 41.09 & 0.12 & 60.8\% \\
MiniMax3 & +Physical & 0.79 & 1.39 & 31.73 & 0.10 & 57.7\% \\
MiniMax3 & +RAG & 0.88 & 1.43 & 21.81 & 0.10 & 85.9\% \\
MiniMax3 & Full & 0.88 & 1.31 & 16.74 & 0.09 & 95.1\% \\
\midrule
GLM-5.3-Flash & Baseline & 0.54 & 3.01 & 124.20 & 0.22 & 42.6\% \\
GLM-5.3-Flash & Physical-only & 0.79 & 1.37 & 28.13 & 0.11 & 60.0\% \\
GLM-5.3-Flash & +Physical & 0.80 & 1.33 & 31.78 & 0.10 & 58.2\% \\
GLM-5.3-Flash & +RAG & 0.87 & 1.23 & 23.73 & 0.09 & 88.6\% \\
GLM-5.3-Flash & Full & 0.88 & 1.32 & 23.68 & 0.08 & 90.1\% \\
\midrule
Qwen3-VL-Plus & Baseline & 0.44 & 4.58 & 116.70 & 0.29 & 26.5\% \\
Qwen3-VL-Plus & Physical-only & 0.65 & 1.46 & 20.45 & 0.08 & 41.2\% \\
Qwen3-VL-Plus & +Physical & 0.80 & 1.80 & 18.09 & 0.06 & 52.4\% \\
Qwen3-VL-Plus & +RAG & 0.93 & 0.60 & 10.02 & 0.04 & 95.1\% \\
Qwen3-VL-Plus & Full & 0.94 & 0.60 & 9.60 & 0.03 & 95.6\% \\
\bottomrule
\end{tabularx}
\end{table}

\paragraph{Physical symbols improve quantitative recognition.}
Table~\ref{tab:real-cross-model} shows that switching from the image-only Baseline to Physical-only increased quality-classification macro-F$_1$ by 0.32, 0.21, 0.25, and 0.21 for Qwen3.5-Plus, MiniMax3, GLM-5.3-Flash, and Qwen3-VL-Plus, respectively. The corresponding reductions in acceleration-current MAE were 2.27, 1.57, 1.64, and 3.12~A. Across all four models, pulse-width MAE decreased by 66.71--109.82~ms and flat-top-count MAE by 0.11--0.21. The direction of improvement was therefore consistent for all quantitative metrics, with Qwen3.5-Plus showing the largest gain in quality classification and Qwen3-VL-Plus the largest gain in current estimation. For the present replay task, directly supplying current, pulse width, flat-top count, and other quantities extracted by deterministic algorithms was more reliable than requiring the models to recover quantitative information from waveform images alone. This comparison evaluates two complete input schemes rather than isolating the causal contribution of a single input change.

\paragraph{Visual gains vary across models.}
The comparison between Physical-only and +Physical reveals clear model dependence in the effect of visual input on quality classification, action direction, and continuous-value errors. For Qwen3-VL-Plus, quality-classification macro-F$_1$ increased by 0.15 and directional agreement by 11.2 percentage points. For MiniMax3 and GLM-5.3-Flash, macro-F$_1$ increased by only 0.01, while directional agreement changed by $-3.1$ and $-1.8$ percentage points, respectively. For Qwen3.5-Plus, macro-F$_1$ decreased by 0.06, whereas directional agreement increased by 1.4 percentage points. Current MAE decreased for MiniMax3 and GLM-5.3-Flash, and pulse-width MAE decreased for MiniMax3 and Qwen3-VL-Plus, with MiniMax3 improving on both quantities. These results indicate that the visual channel primarily contributes global structural information, including flat-top interruption, waveform oscillation, and cross-channel morphology, and that the efficiency with which this information is used depends on the model's visual parsing capability. Physical symbols and visual evidence therefore serve complementary functions: the former supply local quantitative readings, and the latter supply global waveform structure.

\paragraph{Historical retrieval strengthens action reproduction.}
From +Physical to +RAG, mean directional agreement increased by 22.4--42.7 percentage points. For Qwen3.5-Plus, MiniMax3, GLM-5.3-Flash, and Qwen3-VL-Plus, it increased from 55.8\%, 57.7\%, 58.2\%, and 52.4\% to 78.2\%, 85.9\%, 88.6\%, and 95.1\%, respectively. At the same time, pulse-width MAE decreased by 8.05--22.87~ms for all four models; quality-classification macro-F$_1$ increased by 0.07--0.13 for MiniMax3, GLM-5.3-Flash, and Qwen3-VL-Plus and decreased slightly, by 0.02, for Qwen3.5-Plus; and current MAE decreased for three models while increasing slightly, by 0.04~A, for MiniMax3. The largest and most consistent retrieval gain therefore occurred in action-direction reproduction, followed by pulse-width estimation and quality classification. This pattern matches the intended role of retrieval: recent trajectories and historical cases provide cross-shot decision context, while quantitative readings of the current shot continue to be supplied by $\Phi$.

\paragraph{The Full configuration achieves the strongest overall performance.}
From +RAG to Full, mean directional agreement increased further for all four models, but by markedly different amounts: by 7.8 and 9.2 percentage points for Qwen3.5-Plus and MiniMax3, and by only 1.5 and 0.5 percentage points for GLM-5.3-Flash and Qwen3-VL-Plus. Quality-classification macro-F$_1$ changed by $-0.03$ to $+0.01$, and flat-top-count MAE decreased by 0.01 for every model. The human-feedback protocol therefore acted mainly on action direction, yielding larger gains for models whose action reproduction had not saturated under retrieval and limited additional change for models already near the ceiling. Continuous-value errors varied by metric: current MAE decreased further for MiniMax3, remained unchanged for Qwen3-VL-Plus, and increased slightly for Qwen3.5-Plus and GLM-5.3-Flash, whereas pulse-width MAE decreased further for MiniMax3, GLM-5.3-Flash, and Qwen3-VL-Plus. Considering quality classification, continuous-value estimation, complex-waveform recognition, and action-direction reproduction together, Qwen3-VL-Plus Full achieved the strongest overall performance. Its quality-classification macro-F$_1$ was 0.94; its current, pulse-width, and flat-top-count MAEs were 0.60~A, 9.60~ms, and 0.03, respectively; and its mean directional agreement across the five parameters was 95.6\%. This configuration was therefore selected for the subsequent on-device evaluation. Complete results for continuous values, classification, and individual action parameters, together with a visual summary of cross-model module effects, are provided in Appendix~\ref{app:detailed-results}.

\paragraph{Summary of the offline evaluation.}
The results establish a clear division of labor among the modules: physical symbols provide a quantitatively stable baseline across models, visual evidence supplements model-dependent waveform morphology, and historical retrieval substantially strengthens cross-shot action-direction reproduction. This pattern is consistent with the action-neighbor conflict analysis in Section~\ref{sec:drift-result}, indicating that historical trajectories contain decision-relevant information beyond the current-shot observation when the latent device state is not directly observable. The shared trends across the four models further suggest that the principal gains arise from input construction and evidence organization rather than dependence on a single foundation model. The symbolic-grounding, memory-retrieval, and feasibility-check interfaces can therefore remain stable while the foundation model is replaced according to deployment requirements. Section~\ref{sec:online} uses Qwen3-VL-Plus Full for supervised execution on the physical system.

\subsection{Supervised On-Device Execution Validation}\label{sec:online}
To further assess the deployability and operational stability of the proposed method on a physical scientific instrument, the Qwen3-VL-Plus Full configuration, which achieved the best offline evaluation results, was integrated with EXL-50U for supervised shot-to-shot execution. The on-device validation comprised two complementary levels. Long-term operation across multiple ion sources and comparison with historical manual-tuning records were used to assess cross-source engineering reliability. Three consecutive experiments on the same ion source at Test Bench 2 were used to examine whether the system could progressively advance operating parameters toward prespecified targets on the basis of shot-to-shot feedback.

\paragraph{Routine operation: beam-extraction reliability across ion sources.}
A total of 409 shots were physically executed across five ion sources, all of which were included in the analysis. Using whether beam extraction occurred in each shot according to the operational record as the criterion, 355 shots exhibited beam extraction and 54 did not, corresponding to an overall beam-extraction rate of 86.80\%; the source-specific rates ranged from 82.81\% to 91.84\%. Table~\ref{tab:online-source-success} presents the source-specific results. More than 400 shots covered different ion sources, power-supply conditions, and device states, and every source maintained a beam-extraction rate above 80\%, providing on-device evidence of the system's engineering reliability during sustained operation across ion sources.

\begin{table}[H]
  \centering
  \caption{Summary of physical operation results for five ion sources}
  \label{tab:online-source-success}
  \small
  \begin{tabular}{lrrrr}
    \toprule
    Ion source & \shortstack{Executed\\shots} & \shortstack{With beam\\extraction} & \shortstack{Without beam\\extraction} & \shortstack{Beam-extraction\\rate} \\
    \midrule
    Test Bench 2 & 49 & 45 & 4 & 91.84\% \\
    Source 1, Experimental Bench 2 & 64 & 53 & 11 & 82.81\% \\
    Source 2, Experimental Bench 2 & 83 & 69 & 14 & 83.13\% \\
    Source 3, Experimental Bench 2 & 103 & 91 & 12 & 88.35\% \\
    Source 4, Experimental Bench 2 & 110 & 97 & 13 & 88.18\% \\
    \midrule
    \textbf{All} & \textbf{409} & \textbf{355} & \textbf{54} & \textbf{86.80\%} \\
    \bottomrule
  \end{tabular}
\end{table}

\paragraph{Routine operation: comparison with manual tuning.}
To contextualize the beam-extraction rate, we further selected four ion sources from Experimental Bench 2 for which comparable manual records were available. The 360 shots tuned by the supervised agent formed a subset of the 409 records described above; historical records comprised an additional 1695 shots for which operators directly specified the parameters on the same ion sources. The same beam-extraction criterion was applied to both groups. Figure~\ref{fig:manual-vs-ai} presents the source-specific comparison, and complete counts are provided in Appendix Table~\ref{tab:manual-vs-agent}. Overall, the beam-extraction rate was 83.83\% (1421/1695) in the manual-tuning records and 86.11\% (310/360) in the agent-tuning records, an observed difference of $+2.3$ percentage points. The source-specific differences ranged from $-3.7$ to $+6.0$ percentage points.

\begin{figure}[H]
  \centering
  \includegraphics[width=0.92\linewidth]{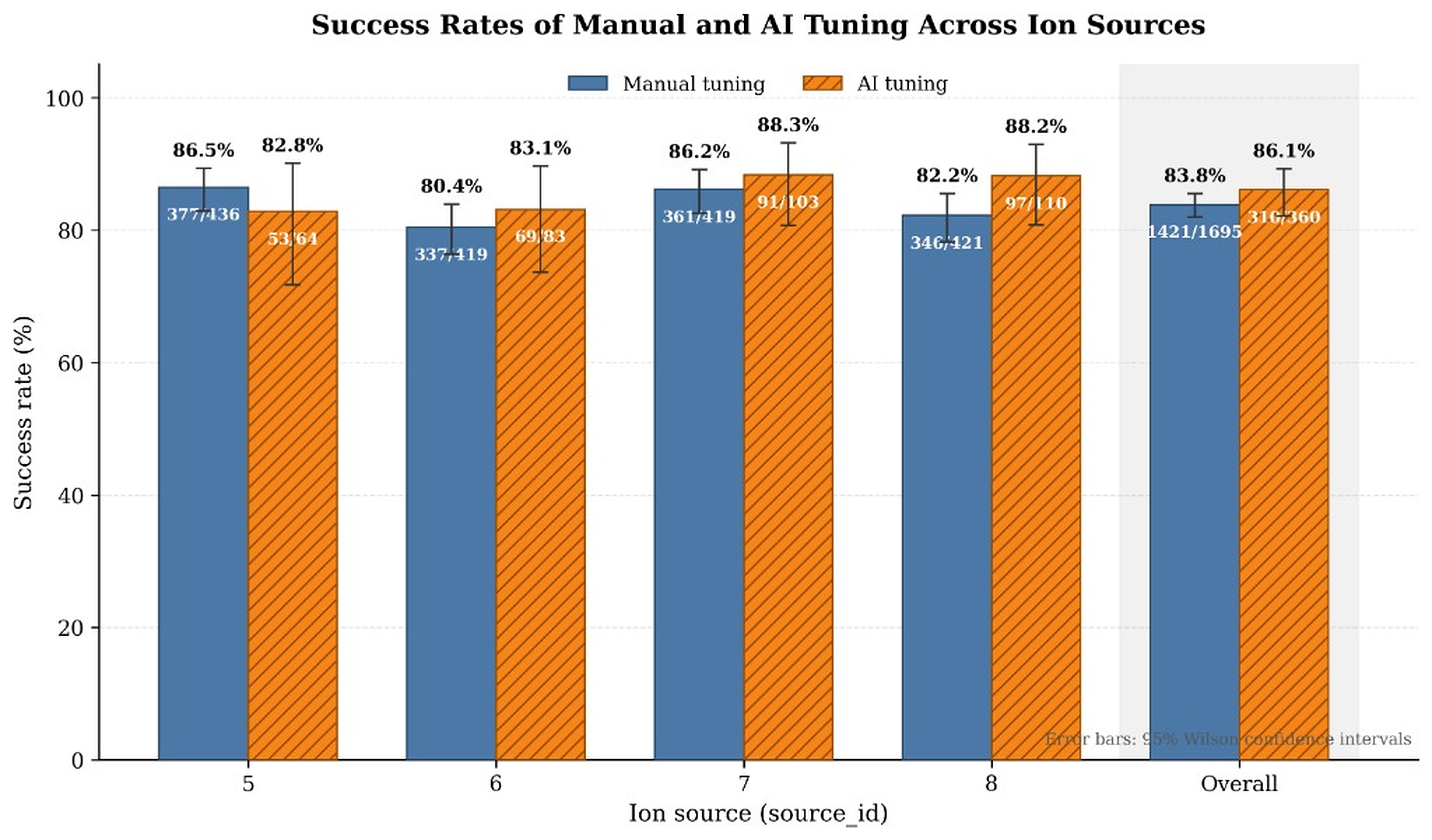}
  \caption{Comparison of shot-level beam-extraction rates between manual tuning and supervised agent tuning (AI tuning in the figure) for four ion sources. Numbers inside the bars denote shots with beam extraction and physically executed shots, and labels above the bars give the corresponding extraction rates. Error bars are 95\% Wilson intervals calculated from shot-level binomial samples and characterize the sampling uncertainty of each beam-extraction rate. Source IDs 5--8 on the horizontal axis are the identifiers in the raw records and correspond, respectively, to Sources 1--4 of Experimental Bench 2 in Table~\ref{tab:online-source-success}.}
  \label{fig:manual-vs-ai}
\end{figure}

The two sets of records cover the same ion sources, diagnostic chain, and operating team, and span different operating periods and task objectives. Under these shared device conditions, agent tuning achieved a beam-extraction rate comparable to that of experienced operators, providing a historical reference for its engineering feasibility. Contemporaneous comparisons with matched task objectives and device states will further quantify the relative performance of the two operating modes.

\paragraph{Parameter progression: target-oriented shot-to-shot improvement.}
Building on the routine-operation validation, we further examined whether operating conditions progressed toward their targets over successive shots using three consecutive experiments on the same ion source at Test Bench 2. All three experiments performed shot-to-shot tuning toward prespecified targets, and the system protocol generated the next-shot recommendation from feedback on the preceding shot. The corresponding shot ranges were 11394--11402, 11884--11891, and 11948--11953, totaling 23 shots; these shots were also a subset of the 49 operational records for that ion source. The available archive provides complete outcome-level trajectories for testing whether the observed responses evolved consistently with progressive parameter adjustment, thereby supporting an outcome-based assessment of target-oriented tuning.

Specifically, a shot was considered to contain a valid beam interval if the absolute acceleration current reached 1~A continuously for at least 10~ms. The duration of the longest continuous interval was taken as the beam pulse width, and the median acceleration voltage and acceleration current within that interval were calculated. This waveform-based criterion serves a different purpose from the record-based criterion used in the multi-source summary: the former constructs auditable shot-to-shot response trajectories, whereas the latter summarizes shot-level beam extraction outcomes in the existing operational records.

\begin{figure}[H]
  \centering
  \includegraphics[width=\linewidth]{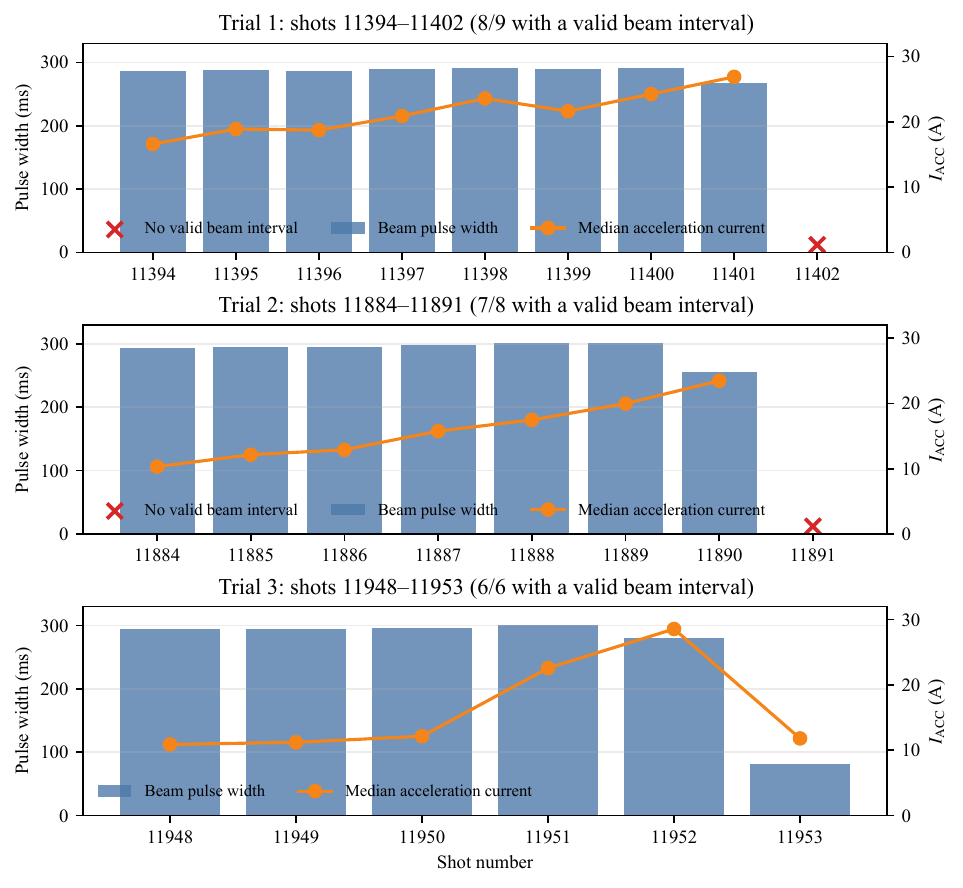}
  \caption{Shot-to-shot response trajectories from three consecutive experiments under the target-oriented protocol. Bars indicate beam pulse width, lines indicate the median acceleration current within the valid interval, and red crosses indicate shots in which no valid beam interval was detected.}
  \label{fig:online}
\end{figure}

Fig.~\ref{fig:online} presents the shot-to-shot trajectories, and Table~\ref{tab:online-three-trials} summarizes the three experiments. Valid beam intervals were detected in 21 of the 23 shots, with these shots spanning an acceleration-voltage range of 21.06--44.54~kV.

\begin{table}[H]
  \centering
  \caption{Waveform statistics from three consecutive experiments on the same ion source at Test Bench 2. Voltage, current, and pulse width are reported as ranges of shot-level statistics among shots with a valid beam interval.}
  \label{tab:online-three-trials}
  \small
  \setlength{\tabcolsep}{4pt}
  \begin{tabularx}{\linewidth}{l*{4}{>{\centering\arraybackslash}X}}
    \toprule
    Metric & Trial 1 & Trial 2 & Trial 3 & Total \\
    \midrule
    Shot range & 11394--11402 & 11884--11891 & 11948--11953 & --- \\
    Number of shots & 9 & 8 & 6 & 23 \\
    Shots with a valid beam interval & 8 & 7 & 6 & 21 \\
    Valid-beam-interval rate & 88.89\% & 87.50\% & 100.00\% & 91.30\% \\
    Median acceleration-voltage range/kV & 26.39--33.36 & 21.06--44.54 & 22.43--38.44 & 21.06--44.54 \\
    Median acceleration-current range/A & 16.60--26.88 & 10.31--23.46 & 10.89--28.56 & 10.31--28.56 \\
    Valid pulse-width range/ms & 267--291 & 256--302 & 82--301 & 82--302 \\
    \bottomrule
  \end{tabularx}
\end{table}

\paragraph{Interpretation and tuning dynamics.}
In Fig.~\ref{fig:online}, the operating parameters in all three experiments exhibited an overall stepwise increase during shot-to-shot tuning. The median acceleration current increased from 16.60~A at shot 11394 to 26.88~A at shot 11401 in the first experiment, from 10.31~A at shot 11884 to 23.46~A at shot 11890 in the second, and from 10.89~A at shot 11948 to 28.56~A at shot 11952 in the third. Shots 11402 and 11891 produced no valid beam interval, while the beam pulse width at shot 11953 decreased to 82~ms; all three observations occurred after the parameters had been advanced to relatively high levels. In the context of the on-site tuning process, these outcomes reflect the greater state-matching demands of operation in the high-parameter regime: after the parameters are increased, multiple rounds of shot-to-shot adjustment are typically required to progressively improve the ion-source state and establish stable beam extraction at the higher operating level. These terminal shots therefore represent transitional tuning states during continued advancement into the high-parameter regime, while all three trajectories retain a clear overall direction of increasing the operating parameters.

Taken together, the multi-source records support supervised integration and routine operation of the system across different ion sources, and the comparison with manual records provides a reference level under the same device conditions. The three consecutive experiments further show that, under the target-oriented protocol, continuous shot-to-shot tuning advanced operation from a lower-parameter regime toward higher voltage and current while allowing continued iterative improvement of the ion-source state at the higher operating level. The response trajectories therefore support the system's target-oriented parameter-advancement capability.

\subsection{General Discussion and Outlook}

The findings collectively support a central interpretation: shot-to-shot tuning of a scientific instrument is continual adaptation to a device state that evolves with operational history. Different quality outcomes among neighboring actions, together with their dependence on inter-shot separation, indicate that history is an essential component for characterizing the current operating condition and organizing the next-shot decision. The central role of the MLLM in this framework is therefore to use traceable cross-shot context to address the information gap created by an unobserved latent state.

The cross-model and on-device results further reveal the complementary roles of the three system components. Deterministic physical symbols answer what occurred in the current shot; history-conditioned memory addresses how the next action should be adjusted in light of the observed trajectory; and the independent feasibility check determines whether the candidate adjustment can be dispatched safely. This separation maintains clear responsibility boundaries among shot-level measurement, cross-shot reasoning, and action execution, and helps explain why the overall pattern remains consistent when the foundation model is changed. During consecutive physical experiments, transient mismatches in the high-parameter regime reflected the need for multiple interactions to condition the ion-source state. Continued advancement toward the target demonstrates that history-conditioned decision-making naturally represents the practical cycle of increasing parameters, observing the response, improving the source state, and advancing again. Operation across multiple sources and comparison with manual records further indicate that the method has progressed from offline behavioral reproduction to supervised physical operation at an engineering-relevant level.

In practical operation, the system has been integrated into the routine operation of the EXL-50U NBI systems: its waveform-recognition results are used continuously in routine operation, and it assisted in achieving the 2.5~MW injection power of the two positive-ion NBI systems in 2025. This indicates that multimodal agents can share the interpretation and decision-making workload of experimental operation on large scientific facilities.

The broader significance of this work extends beyond ion-source operation. For scientific instruments that lack high-fidelity simulators, exhibit history-dependent latent-state evolution, and impose high costs per experiment, deterministic measurement, history-conditioned decision-making, and action constraints can be organized as modular interfaces. This organization permits foundation models to be replaced or upgraded while preserving a traceable evidence chain and the existing protection hierarchy of the instrument. The next stage can progress from engineering feasibility to quantitative performance gains and higher-autonomy operation through unified logging of shot-level targets, candidate actions, check outcomes, executed actions, human interventions, and waveform outcomes. Contemporaneous comparisons with matched tasks and device states, together with cross-device evaluation, expanded decrease-action samples, shots to first target attainment, abnormal-event rates, and independent retrieval ablations, will enable systematic characterization of sample efficiency, transferability, and closed-loop tuning benefit.

\section{Conclusion}

Scientific-instrument responses can evolve with latent operating conditions and operational history, making tuning strategies based on a static action--response mapping unreliable across successive operations. Taking NBI ion-source tuning as a representative setting, this study develops a multimodal agent for adaptive tuning under nonstationary responses. Deterministic signal processing converts raw waveforms into auditable physical symbols and visual evidence; a dual-layer memory conditions next-shot decisions on within-task trajectories, cross-task experience, and procedural knowledge; and an independent feasibility layer checks candidate actions before dispatch. This separation assigns evidence integration and candidate-action generation to the MLLM while retaining quantitative measurement and final protection within deterministic algorithms, the PLC, and hardware interlocks.

The experimental results support the framework at three levels. First, the quality-outcome conflict rate was 54.7\% among action-neighbor shot pairs and remained 44.1\% for pairs with identical parameters; its increase with inter-shot separation demonstrates the importance of operational history to the device response. Second, across 455 matched records from four platforms, the grounding operator achieved pooled MAEs of 0.68~A for acceleration current and 10.42~ms for main flat-top width, with event-count F$_1$ values of 97.61\% for flat tops and 99.66\% for breakdowns. In strict chronological replay of the same records, physical symbols increased quality-classification macro-F$_1$ by 0.21--0.32 relative to waveform images alone across four models, while retrieved context increased mean directional agreement by 22.4--42.7 percentage points. These results verify the complementary roles of reliable physical representation and history-conditioned decision-making. Third, the system completed 409 shots of physical execution across five ion sources with an overall beam-extraction rate of 86.80\%. On four comparable sources, the beam-extraction rates were 86.11\% for supervised agent tuning and 83.83\% for manual-tuning records. All three consecutive experiments advanced the operating parameters toward higher levels, demonstrating target-oriented progression and continued source conditioning in the high-parameter regime.

This work therefore establishes a simulator-free route for tuning scientific instruments with nonstationary responses by combining history-conditioned decision-making with deterministic measurement and constrained execution. Its central contribution is to embed a multimodal model within an auditable, replaceable, and protection-compatible shot-to-shot decision architecture, in which the model works alongside deterministic physical measurement and the existing safety controls. The resulting framework provides a transferable methodological basis for other scientific instruments characterized by history dependence, partial observability, and expensive experimentation, while creating the experimental and engineering foundation for subsequent quantification of closed-loop benefit, sample efficiency, and cross-device generalization.

\section*{Acknowledgements}
This work was supported by Beijing ENN Fusion Energy Science and Technology Co., Ltd. The authors thank the EXL-50U team and the NBI operation team for their support in device operation, data acquisition, and on-device experiments.

\appendix
\makeatletter
\renewcommand{\@seccntformat}[1]{Appendix~\csname the#1\endcsname\quad}
\makeatother
\section{Detailed Cross-Model Results}\label{app:detailed-results}

\begin{figure}[H]
  \centering
  \includegraphics[width=\linewidth]{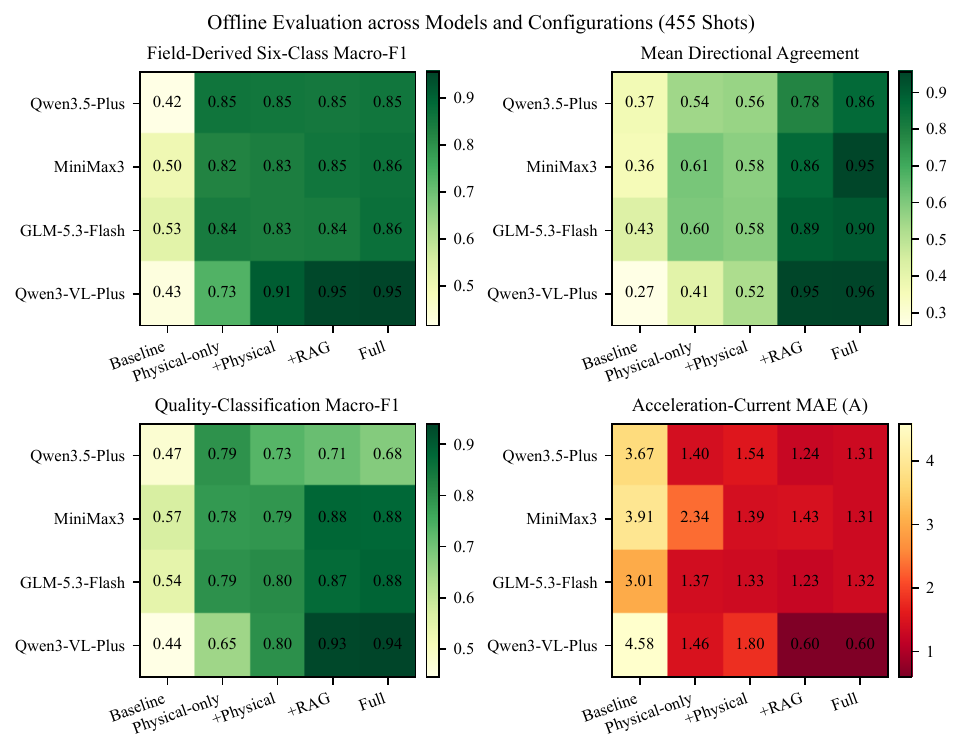}
  \caption{Offline performance of four models under five configurations. From left to right, the panels report field-derived six-class macro-F$_1$, mean directional agreement across five action parameters, quality-classification macro-F$_1$, and acceleration-current MAE.}
  \label{fig:real-cross-model}
\end{figure}

Table~\ref{tab:real-cross-model} in the main text reports the aggregate results. This
appendix provides the complete disaggregated results for continuous-variable errors,
quality classification, and the five action dimensions, together with
Fig.~\ref{fig:real-cross-model} as a visual summary of module effects across models
and configurations.

\begin{table}[H]
\centering
\caption{Detailed recognition results for the four models. Continuous variables are reported as MAE/RMSE; quality classification is reported as Accuracy and macro-F$_1$.}
\label{tab:recognition-detail-all}
\small
\setlength{\tabcolsep}{3pt}
\begin{tabularx}{\linewidth}{ll*{8}{>{\centering\arraybackslash}X}}
\toprule
Model & Configuration & \multicolumn{2}{c}{Current/A} & \multicolumn{2}{c}{Pulse width/ms} & \multicolumn{2}{c}{\shortstack{Flat-top\\count}} & \multicolumn{2}{c}{\shortstack{Quality\\classification}} \\
\cmidrule(lr){3-4}\cmidrule(lr){5-6}\cmidrule(lr){7-8}\cmidrule(lr){9-10}
 & & MAE & RMSE & MAE & RMSE & MAE & RMSE & Accuracy & F$_1$ \\
\midrule
Qwen3.5-Plus & Baseline & 3.67 & 14.41 & 138.19 & 328.19 & 0.27 & 0.66 & 0.51 & 0.47 \\
Qwen3.5-Plus & Physical-only & 1.40 & 5.13 & 28.37 & 86.58 & 0.09 & 0.41 & 0.79 & 0.79 \\
Qwen3.5-Plus & +Physical & 1.54 & 7.08 & 42.23 & 121.89 & 0.09 & 0.34 & 0.73 & 0.73 \\
Qwen3.5-Plus & +RAG & 1.24 & 4.92 & 19.36 & 67.51 & 0.09 & 0.37 & 0.74 & 0.71 \\
Qwen3.5-Plus & Full & 1.31 & 5.05 & 24.33 & 77.34 & 0.08 & 0.36 & 0.84 & 0.68 \\
\midrule
MiniMax3 & Baseline & 3.91 & 14.89 & 107.80 & 277.65 & 0.24 & 0.62 & 0.62 & 0.57 \\
MiniMax3 & Physical-only & 2.34 & 7.08 & 41.09 & 103.44 & 0.12 & 0.43 & 0.77 & 0.78 \\
MiniMax3 & +Physical & 1.39 & 5.24 & 31.73 & 91.71 & 0.10 & 0.38 & 0.79 & 0.79 \\
MiniMax3 & +RAG & 1.43 & 5.27 & 21.81 & 71.18 & 0.10 & 0.41 & 0.88 & 0.88 \\
MiniMax3 & Full & 1.31 & 5.07 & 16.74 & 62.43 & 0.09 & 0.39 & 0.88 & 0.88 \\
\midrule
GLM-5.3-Flash & Baseline & 3.01 & 11.91 & 124.20 & 353.68 & 0.22 & 0.59 & 0.59 & 0.54 \\
GLM-5.3-Flash & Physical-only & 1.37 & 5.21 & 28.13 & 84.30 & 0.11 & 0.44 & 0.79 & 0.79 \\
GLM-5.3-Flash & +Physical & 1.33 & 5.11 & 31.78 & 96.92 & 0.10 & 0.37 & 0.80 & 0.80 \\
GLM-5.3-Flash & +RAG & 1.23 & 4.93 & 23.73 & 79.51 & 0.09 & 0.38 & 0.87 & 0.87 \\
GLM-5.3-Flash & Full & 1.32 & 5.14 & 23.68 & 74.29 & 0.08 & 0.36 & 0.88 & 0.88 \\
\midrule
Qwen3-VL-Plus & Baseline & 4.58 & 15.41 & 116.70 & 219.96 & 0.29 & 0.70 & 0.48 & 0.44 \\
Qwen3-VL-Plus & Physical-only & 1.46 & 5.08 & 20.45 & 67.70 & 0.08 & 0.36 & 0.69 & 0.65 \\
Qwen3-VL-Plus & +Physical & 1.80 & 7.44 & 18.09 & 71.06 & 0.06 & 0.32 & 0.80 & 0.80 \\
Qwen3-VL-Plus & +RAG & 0.60 & 3.36 & 10.02 & 37.49 & 0.04 & 0.27 & 0.93 & 0.93 \\
Qwen3-VL-Plus & Full & 0.60 & 3.36 & 9.60 & 37.29 & 0.03 & 0.27 & 0.94 & 0.94 \\
\bottomrule
\end{tabularx}
\end{table}

\begin{table}[H]
\centering
\caption{Directional agreement (Accuracy) for five parameters across the four models. The mean is the arithmetic mean of the directional agreement for the five parameters.}
\label{tab:action-detail-all}
\small
\setlength{\tabcolsep}{3pt}
\begin{tabularx}{\linewidth}{ll*{6}{>{\centering\arraybackslash}X}}
\toprule
Model & Configuration & $V_{\rm ACC}$ & $V_{\rm FIL}$ & $V_{\rm ARC}$ & $Q$ & $V_{\rm DEC}$ & Mean \\
\midrule
Qwen3.5-Plus & Baseline & 55.6\% & 47.9\% & 26.6\% & 18.7\% & 34.9\% & 36.7\% \\
Qwen3.5-Plus & Physical-only & 66.6\% & 69.2\% & 59.6\% & 39.8\% & 36.9\% & 54.4\% \\
Qwen3.5-Plus & +Physical & 66.6\% & 76.9\% & 56.9\% & 40.9\% & 37.8\% & 55.8\% \\
Qwen3.5-Plus & +RAG & 87.9\% & 82.2\% & 72.3\% & 75.4\% & 73.2\% & 78.2\% \\
Qwen3.5-Plus & Full & 89.7\% & 84.4\% & 85.1\% & 81.3\% & 89.7\% & 86.0\% \\
\midrule
MiniMax3 & Baseline & 49.2\% & 48.8\% & 23.7\% & 19.1\% & 37.4\% & 35.6\% \\
MiniMax3 & Physical-only & 73.2\% & 73.6\% & 60.7\% & 48.4\% & 48.4\% & 60.8\% \\
MiniMax3 & +Physical & 66.8\% & 73.6\% & 56.0\% & 45.7\% & 46.4\% & 57.7\% \\
MiniMax3 & +RAG & 91.0\% & 89.2\% & 88.1\% & 81.3\% & 80.0\% & 85.9\% \\
MiniMax3 & Full & 95.6\% & 95.4\% & 94.7\% & 94.1\% & 95.8\% & 95.1\% \\
\midrule
GLM-5.3-Flash & Baseline & 61.3\% & 65.9\% & 19.8\% & 16.9\% & 49.0\% & 42.6\% \\
GLM-5.3-Flash & Physical-only & 77.6\% & 69.0\% & 59.3\% & 44.4\% & 49.5\% & 60.0\% \\
GLM-5.3-Flash & +Physical & 71.0\% & 70.5\% & 56.0\% & 44.8\% & 48.6\% & 58.2\% \\
GLM-5.3-Flash & +RAG & 92.7\% & 87.7\% & 88.6\% & 88.6\% & 85.3\% & 88.6\% \\
GLM-5.3-Flash & Full & 93.8\% & 88.1\% & 89.7\% & 88.1\% & 90.5\% & 90.1\% \\
\midrule
Qwen3-VL-Plus & Baseline & 23.1\% & 41.3\% & 19.1\% & 18.9\% & 30.1\% & 26.5\% \\
Qwen3-VL-Plus & Physical-only & 56.7\% & 37.4\% & 38.9\% & 38.5\% & 34.7\% & 41.2\% \\
Qwen3-VL-Plus & +Physical & 69.2\% & 74.1\% & 47.9\% & 31.6\% & 39.1\% & 52.4\% \\
Qwen3-VL-Plus & +RAG & 97.8\% & 95.6\% & 94.3\% & 95.8\% & 91.9\% & 95.1\% \\
Qwen3-VL-Plus & Full & 98.0\% & 96.0\% & 94.9\% & 96.3\% & 92.5\% & 95.6\% \\
\bottomrule
\end{tabularx}
\end{table}

\section{Detailed Results for Manual and Agent Tuning}

Figure~\ref{fig:manual-vs-ai} in the main text visually compares the shot-level beam-extraction
rates of manual tuning and supervised agent tuning.
Table~\ref{tab:manual-vs-agent} reports the number of shots with beam extraction, the number of
executed shots, and the difference in beam-extraction rate for each ion source.

\begin{table}[H]
  \centering
  \caption{Comparison of shot-level beam-extraction rates between manual tuning and supervised agent tuning on four ion sources}
  \label{tab:manual-vs-agent}
  \small
  \setlength{\tabcolsep}{3pt}
  \begin{tabularx}{\linewidth}{l*{5}{>{\centering\arraybackslash}X}}
    \toprule
    \multirow{2}{*}{Ion source}
      & \multicolumn{2}{c}{Manual tuning}
      & \multicolumn{2}{c}{Agent tuning}
      & \multirow{2}{*}{\shortstack{Difference\\(percentage points)}} \\
    \cmidrule(lr){2-3}\cmidrule(lr){4-5}
      & \shortstack{Beam extracted/\\executed} & \shortstack{Extraction\\rate} & \shortstack{Beam extracted/\\executed} & \shortstack{Extraction\\rate} & \\
    \midrule
    ID 5 (Source 1) & 377/436 & 86.47\% & 53/64  & 82.81\% & $-3.7$ \\
    ID 6 (Source 2) & 337/419 & 80.43\% & 69/83  & 83.13\% & $+2.7$ \\
    ID 7 (Source 3) & 361/419 & 86.16\% & 91/103 & 88.35\% & $+2.2$ \\
    ID 8 (Source 4) & 346/421 & 82.19\% & 97/110 & 88.18\% & $+6.0$ \\
    \midrule
    \textbf{Total} & \textbf{1421/1695} & \textbf{83.83\%}
      & \textbf{310/360} & \textbf{86.11\%} & $\mathbf{+2.3}$ \\
    \bottomrule
  \end{tabularx}
\end{table}

\section{Control Parameters and Target Thresholds}

Table~\ref{tab:parameter-list} lists the five parameters directly adjusted by the agent in
this study. The action bounds use the permissible ranges in the PLC configuration rather
than the extrema in the historical samples, because the historical data cover a narrower
range than the device's permissible operating range.

The operating procedures also specify recommended per-shot adjustment steps for each
control variable. For acceleration voltage, filament voltage, arc voltage, and gas flow,
the intervals in the table denote the steps normally used when a parameter is adjusted.
The upper bound is used as the per-shot change limit in Eq.~\eqref{eq:feasible}; an
unchanged parameter is not subject to the lower bound of the interval. Suppression voltage
is output as the fifth candidate action variable. It does not use an independent fixed
step size but is constrained during the action feasibility check to 4\%--10\% of the
acceleration voltage. The ``arc-current power-supply adjustment'' in the original
procedures corresponds to the arc-voltage setting $V_{\mathrm{ARC}}$ in the action space
used here.

\begin{table}[H]
  \centering
  \caption{Agent action parameters and PLC permissible constraints}
  \label{tab:parameter-list}
  \small
  \begin{tabularx}{\linewidth}{|l|l|c|>{\raggedright\arraybackslash}X|>{\raggedright\arraybackslash}X|}
    \hline
    Symbol & Physical quantity & Unit & PLC permissible range & Maximum per-shot change \\ \hline
    $V_{\mathrm{ACC}}$ & Acceleration voltage & kV
      & $5$--$45$
      & Adjustment step: 1--5~kV \\ \hline
    $V_{\mathrm{FIL}}$ & Filament voltage & V
      & $5$--$12$
      & Adjustment step: 0.1--0.5~V \\ \hline
    $V_{\mathrm{ARC}}$ & Arc voltage & V
      & $70$--$130$
      & Adjustment step: 1--5~V \\ \hline
    $Q$ & Gas flow & sccm
      & $100$--$700$
      & Adjustment step: 10--50~sccm \\ \hline
    $V_{\mathrm{DEC}}$ & Suppression voltage & V
      & $0$--$3200$
      & Constrained during the check to 4\%--10\% of $V_{\mathrm{ACC}}$ \\ \hline
  \end{tabularx}
\end{table}

\section{Signal-Processing Thresholds}

Table~\ref{tab:signal-thresholds} lists the signal-processing thresholds actually used by
the grounding operator $\Phi$. The absolute acceleration-current amplitude threshold is
$1$~A for all four platforms. Pulse selection and flat-top localization use relative
coefficients $r_{\mathrm{sel}}=0.8$ and $r_{\mathrm{flat}}=0.7$, respectively. Breakdown
detection uses a single current threshold, a short time window, and a voltage condition.

\begin{table}[H]
  \centering
  \caption{Signal-processing thresholds of the grounding operator $\Phi$}
  \label{tab:signal-thresholds}
  \small
  \begin{tabularx}{\linewidth}{|l|>{\raggedright\arraybackslash}p{3.6cm}|c|>{\raggedright\arraybackslash}X|}
    \hline
    Symbol & Meaning & Unit & Value and determination method \\ \hline
    $w_{\mathrm{med}}$ & Median-filter window width & samples
      & $15$ \\ \hline
    $3\sigma$ & Measurement-spike rejection bound & dimensionless
      & $3$; $\lvert x-\mathrm{medfilt}\rvert$ is compared with the global standard deviation, rather than a local baseline \\ \hline
    $\alpha$ & Fraction trimmed from each end for the trimmed mean & \%
      & $30$ \\ \hline
    $w_{\mathrm{trim}}$ & Sliding-window width for the trimmed mean & samples
      & $21$ \\ \hline
    $\theta_{\mathrm{abs}}$ & Absolute-amplitude threshold for candidate pulses & A
      & $1$; the acceleration-current threshold is identical across all four platforms \\ \hline
    $r_{\mathrm{sel}}$ & Relative coefficient for candidate pulses & dimensionless
      & $0.8$; relative to the maximum pulse mean \\ \hline
    $r_{\mathrm{flat}}$ & Relative coefficient for flat-top amplitude & dimensionless
      & $0.7$; relative to the smoothed peak \\ \hline
    $\theta_{\nabla}$ & Gradient threshold for flat-top boundaries & A/sample (approximately A/ms)
      & $1$; the gradient is taken with respect to the sample index, corresponding to approximately $1$~A/ms at a sampling interval of $1$~ms \\ \hline
    $\tau_{\min}$ & Minimum valid flat-top duration & samples / ms
      & $5$; compared with the difference in sample indices, corresponding to approximately $5$~ms at 1~kHz; pulses shorter than $10$ samples are also ignored \\ \hline
    $\Delta t_{\mathrm{bd}}$ & Time window for breakdown-induced abrupt-drop detection & ms
      & $10$; the current high-level duration must be ${<}\,10$~ms \\ \hline
    $I_{\mathrm{hi}}$ & Breakdown current threshold & A
      & $2.0$; the same threshold is used for the return criterion \\ \hline
    $V_{\mathrm{off}}$ & Offset for the breakdown low-voltage threshold & kV
      & $5$; the low-level threshold is $V_1+5$, where $V_1$ is the mean voltage during the first $0.1$~s of the record \\ \hline
    $\Delta t_{\mathrm{off}}$ & Extension of the breakdown voltage-detection interval & ms
      & $50$; the voltage-detection window extends from the upward current-threshold crossing to $50$~ms after the current returns below the threshold \\ \hline
    $\rho_{\mathrm{osc}}$ & Prominence threshold for inverted peaks & ---
      & $0.25\times$ the peak-to-peak amplitude, rather than a fixed ampere value \\ \hline
    $w_{\mathrm{osc}}^{\max}$ & Upper width bound for oscillatory dips & samples
      & $50$; no lower bound is imposed \\ \hline
    $\Delta d_{\mathrm{osc}}$ & Depth-change threshold used to exclude breakdown & A
      & $1$; the absolute difference between a dip and the mean of the subsequent $5$ samples must be ${>}\,1$ \\ \hline
    $n_{\mathrm{dip}}^{\min}$ & Minimum number of dips for oscillation detection & count / analysis window
      & $5$; determined over the full acceleration-current segment \\ \hline
  \end{tabularx}
\end{table}

\end{document}